\pdfoutput=1
\ifdefined\pdfobjcompresslevel\pdfobjcompresslevel=0\fi 
\documentclass[11pt]{article}

\usepackage[final]{acl}
\usepackage{booktabs}
\usepackage{tabularx}
\usepackage{times}
\usepackage{latexsym}
\usepackage{algorithmic}
\usepackage{algorithm}
\usepackage{xfp}
\usepackage{multirow}
\usepackage{tikz}
\usetikzlibrary{positioning,arrows.meta,fit,calc}
\usepackage{booktabs}
\usepackage{placeins}

\usepackage{colortbl}
\usepackage{xcolor}
\usepackage{amsmath}
\usepackage{amssymb}
\usepackage{booktabs}
\usepackage{xcolor}

\usepackage{makecell}

\definecolor{lightred}{RGB}{255,235,238}
\definecolor{lightblue}{RGB}{232,244,253}

\newcommand{\hdr}{\ttfamily\bfseries\footnotesize}

\definecolor{blue0}{RGB}{255,255,255}   
\definecolor{blue1}{RGB}{240,247,255}
\definecolor{blue2}{RGB}{228,240,252}
\definecolor{blue3}{RGB}{214,232,248}
\definecolor{blue4}{RGB}{198,224,244}   

\definecolor{pink0}{RGB}{255,255,255}   
\definecolor{pink1}{RGB}{252,238,242}
\definecolor{pink2}{RGB}{248,226,232}
\definecolor{pink3}{RGB}{244,214,222}
\definecolor{pink4}{RGB}{240,200,212}   

\definecolor{g1}{RGB}{245, 252, 248}   
\definecolor{g2}{RGB}{232, 246, 238}
\definecolor{g3}{RGB}{216, 240, 226}
\definecolor{g4}{RGB}{196, 232, 212}
\definecolor{g5}{RGB}{176, 224, 198}  

\newcommand{\pink}[2]{\cellcolor{pink#1}{#2}}

\usepackage[T1]{fontenc}

\usepackage[utf8]{inputenc}
\usepackage{iftex}
\ifPDFTeX
  \usepackage[safe]{tipa}
  \usepackage{pifont}
  \DeclareUnicodeCharacter{028A}{\textipa{\textupsilon}}
  \DeclareUnicodeCharacter{0279}{\textipa{\textturnr}}
  \DeclareUnicodeCharacter{026A}{\textipa{\textsci}}
  \DeclareUnicodeCharacter{025C}{\textipa{\textrevepsilon}}
  \DeclareUnicodeCharacter{02DE}{\textipa{\textrhoticity}}
  \DeclareUnicodeCharacter{0261}{\textipa{\textscriptg}}
  \DeclareUnicodeCharacter{028C}{\textipa{\textturnv}}
  \DeclareUnicodeCharacter{0292}{\textipa{\textyogh}}
  \DeclareUnicodeCharacter{025B}{\textipa{\textepsilon}}
  \DeclareUnicodeCharacter{0251}{\textipa{\textscripta}}
  \DeclareUnicodeCharacter{274C}{\ding{56}}
  \DeclareUnicodeCharacter{2713}{\ding{51}}
  \DeclareUnicodeCharacter{2717}{\ding{55}}
  \DeclareUnicodeCharacter{26A0}{\ensuremath{\triangle}}
\fi

\usepackage{microtype}

\usepackage{inconsolata}

\usepackage{graphicx}
\graphicspath{{latex/}{./}} 

\title{A Benchmark for Controllable Speaking-Style Captioning in Audio-Language Models}

\author{
 \textbf{Shreeram Suresh Chandra\textsuperscript{1,2*,3}},
 \textbf{Ashwyn Sharma\textsuperscript{2}},
 \textbf{Liang Lu\textsuperscript{2}},
  \textbf{Xiutian Zhao\textsuperscript{1}},
 \textbf{Leda Sari\textsuperscript{2}}, \\
 \textbf{Bj\"orn Schuller \textsuperscript{4}},
  \textbf{Berrak Sisman\textsuperscript{1}}
\\
 \textsuperscript{1}Center for Language and Speech Processing, Johns Hopkins University, USA \\
 \textsuperscript{2}Otter AI \hspace{0.5cm}
 \textsuperscript{3}The University of Texas at Dallas, USA \\
 \textsuperscript{4}GLAM -- Group on Language, Audio, \& Music, Imperial College London, UK \\
 {\small \textsuperscript{*}Work done during an internship at Otter AI.}
}

\newcommand{\tealc}[1]{#1}
\begin{document}
\maketitle
\begin{abstract}

Speaking-style captioning (SSC) aims to generate natural language descriptions of \textit{how} speech is delivered, capturing paralinguistic attributes such as vocal timbre, prosody, and expressivity. Many downstream applications, including conversational AI agents, controllable speech generation, and large-scale audio annotation, require dimension-specific style captions that describe targeted aspects of speech (e.g., speaker traits, emotion, or delivery style), rather than a single undifferentiated description.
However, existing work lacks a unified formulation for controllable speaking-style captioning. We address this gap by formulating SSC as an instruction-following audio-language task, where explicit instructions specify the speaking-style dimensions to be described. We introduce \textit{StyleInstructCaps}, the first standardized benchmark for controllable speaking-style captioning in audio-language models. \textit{StyleInstructCaps} provides a dataset and evaluation framework that measures metadata groundedness, hallucination, instruction-following ability, generalization to unseen instructions and audio datasets, and speaker-style consistency. Across nine state-of-the-art audio-language models, we find that strong instruction-following ability does not necessarily imply grounded speaking-style understanding, and that SSC-specific instruction-diverse training substantially improves groundedness, controllability, and speaker consistency.\footnote{Code, data, and benchmark are available at \url{https://styleinstructcaps.github.io}}




\end{abstract}
\section{Introduction}

Speaking-style captioning (SSC) has recently emerged as a promising direction for generating natural-language descriptions of how speech is delivered, capturing properties such as vocal timbre, prosody, rhythm, emphasis, emotion, and other paralinguistic cues \cite{yamauchi2024stylecap}. Despite growing interest, SSC still lacks a unified task formulation and evaluation paradigm.

Researchers in audio understanding often treat SSC as an extension of speech comprehension, emphasizing communicative intent and discourse-level meaning \cite{jing2024paraclap,chandra25_interspeech,zhang2025sse}. In contrast, speech synthesis and generative modeling communities typically view speaking-style captions as controllable descriptions of vocal characteristics for expressive text-to-speech \cite{wang2025capspeech,guo2023prompttts} and voice conversion systems \cite{kuan2023towards,hai2024dreamvoice}. These differing perspectives reflect a deeper challenge: \textbf{speaking-style is inherently subjective} \cite{adigwe2024people}, spanning multiple interacting paralinguistic factors such as speaker identity, affect, prosody, and contextual delivery. As a result, existing SSC systems often capture different notions of style, making evaluation and comparison across models difficult.


\begin{figure}[!t]
        \scalebox{1}{
        \includegraphics[draft=false,width=\linewidth]{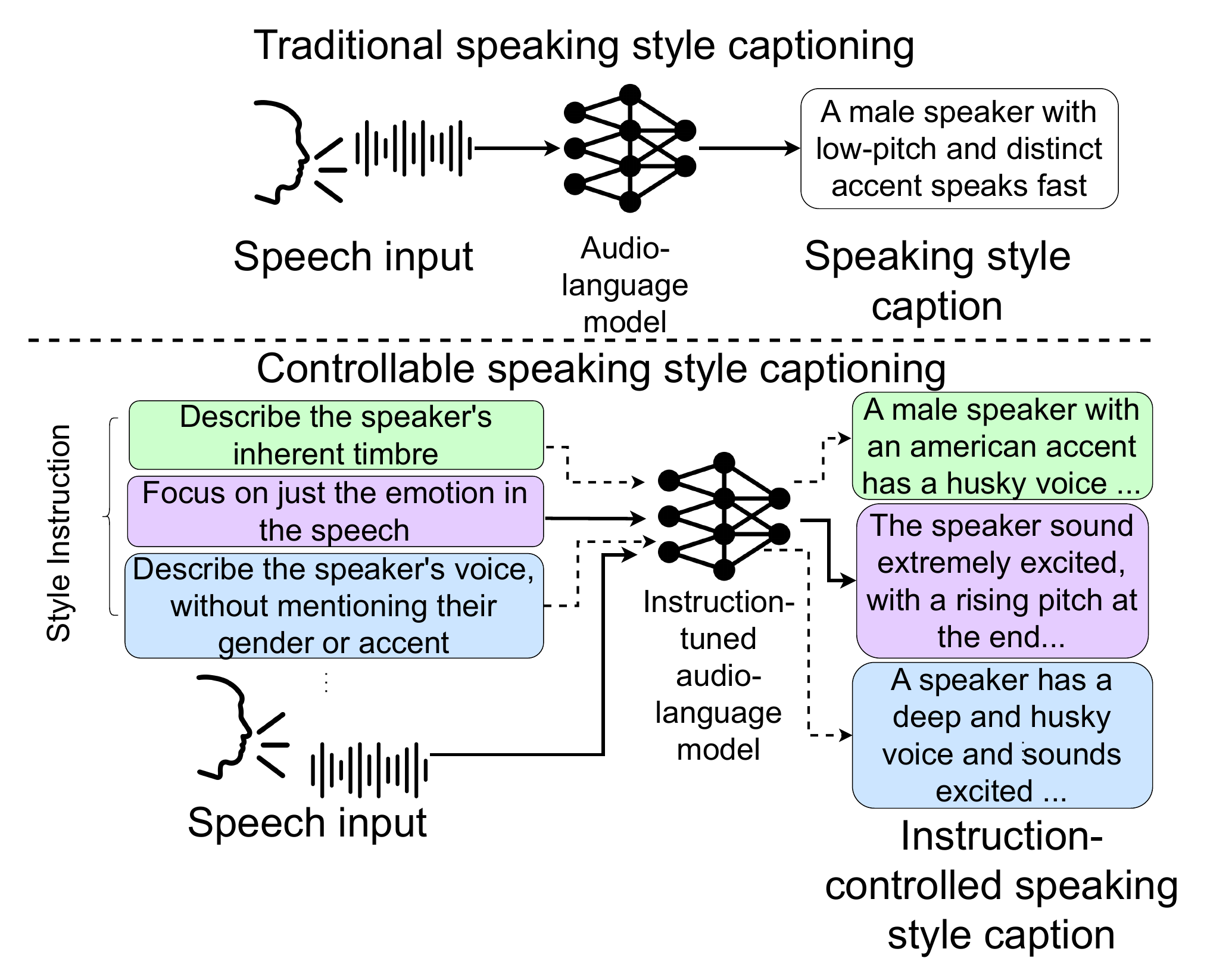}}
        \vspace{-8mm}
        \caption{Speaking-style captioning as an instruction-following audio-language task}
        \label{fig:first}
        \vspace{-4mm}
\end{figure} 


To mitigate this ambiguity, we reformulate SSC as an \textbf{instruction-following audio-language task}, where explicit instructions specify which aspect of speaking-style the model should describe, as illustrated in Figure~\ref{fig:first}. Conventional SSC approaches typically produce a single holistic description of speech, often entangling speaker traits, emotion, prosody, and delivery characteristics, making captions difficult to interpret or evaluate systematically. In contrast, controllable SSC conditions generation on explicit style instructions, enabling the same utterance to yield multiple targeted and complementary descriptions depending on the task objective.

Despite their effectiveness on narrowly defined objectives, existing SSC models often adopt a single view of speaking-style, typically focusing on speaker-centric or emotion-centric attributes, which limits generalization to broader stylistic phenomena. Since speaking-style depends on downstream context, SSC models must be controllable and generate captions conditioned on explicit style instructions. Motivated by this perspective, we introduce \textit{StyleInstructCaps}, the first standardized benchmark for controllable SSC. \textit{StyleInstructCaps} evaluates generated captions along five dimensions:  (1) \textbf{metadata groundedness}, (2) \textbf{hallucination}, (3) \textbf{instruction-following ability}, (4) \textbf{generalization to unseen speech data and prompts}, and (5) \textbf{speaker-style consistency}. 


To support this benchmark, we construct a multi-style speaking-style caption corpus built on top of the ParaSpeechCaps (PSC) dataset \cite{diwan-etal-2025-scaling}. The dataset contains 2.7k hours of speech, where each utterance is paired with six complementary caption types targeting distinct dimensions of speaking-style, including speaker identity, contextual delivery, emotion, pragmatics, listener perception, and holistic style summaries. These parallel annotations enable controlled evaluation of whether models can isolate specific stylistic dimensions, integrate multiple cues when required, and generalize across diverse speaking conditions.

Overall, this work makes three main contributions: (1) We present the first formulation of controllable SSC as an instruction-following audio-language task, in which style descriptions are generated conditioned on explicit stylistic instructions; (2) We introduce StyleInstructCaps, a standardized and challenging benchmark supported by a 2.7k-hour multi-style speaking-style caption corpus with six parallel caption types per utterance;  (3) We evaluate nine state-of-the-art audio-language models, including Audio-Flamingo-3 \cite{goel2025audioflamingo3advancing}, Voxtral \cite{liu2025voxtral}, Qwen2-Audio \cite{Qwen2-Audio}, Qwen3-Omni \cite{xu2025qwen3}, MOSS-Audio \cite{gong2026moss}, Audio Flamingo Next \cite{ghosh2026audio}, MERaLiON-AudioLLM \cite{he2024meralionaudiollmtechnicalreport}, and SALMONN \cite{tang2023salmonn}, together with SSC-specific instruction-tuned variants of SALMONN. We find a consistent gap between instruction-following ability and grounded speaking-style understanding: models often generate stylistically plausible captions while hallucinating unsupported paralinguistic attributes, whereas SSC-specific, instruction-diverse training substantially improves groundedness, controllability, and speaker consistency.

Through this work, we aim to establish a unified foundation for controllable SSC and enable more faithful and interpretable speaking-style understanding in audio-language models.

\section{Related Work}
\subsection{Speaking-Style Captioning for Paralinguistic Speech Understanding}


Speaking-style captioning aims to generate natural-language descriptions of \emph{how} speech is delivered, capturing attributes such as vocal timbre, prosody, emotion, and delivery style \cite{yamauchi2024stylecap}. Unlike categorical paralinguistic tasks such as speaker recognition \cite{hansen2015speaker} and speech emotion recognition \cite{11474737}, which assign a fixed label, and content-oriented tasks such as speech question answering \cite{11278041} and speech summarization \cite{shang24_interspeech,retkowski2025summarizing}, which address what is said rather than how it is said, SSC produces open-ended descriptions that integrate multiple perceptual cues. This capability has become increasingly relevant in large audio-language models, where natural-language descriptions provide a flexible and interpretable interface for expressing subjective speaking-style phenomena \cite{11464563,dhamyal2024prompting,lu2024desta,11447408}. In the reverse direction, speech-generative audio-language models increasingly accept natural-language instructions to produce expressive speech, yet reliably realizing a requested emotion without degrading linguistic content remains challenging \citep{zhang2026traceevctextguidedrelativeaffective, zhao2026neuronlevelemotioncontrolspeechgenerative}.

\subsection{Datasets and Benchmarks}

Progress in SSC depends on (1) speech corpora with rich speaking-style supervision and (2) evaluation protocols that assess generation quality beyond lexical overlap. Existing resources span multiple communities: expressive TTS and style-prompted speech datasets provide speaker and delivery annotations, while paralinguistic benchmarks evaluate reasoning over cues such as emotion, prosody, and conversational pragmatics. Appendix~\ref{app:existing_benchmarks}, Table~\ref{tab:dataset_summary} and Table~\ref{tab:paralinguistic_benchmarks} summarize representative datasets and benchmarks in this space. However, most existing benchmarks emphasize \emph{recognition} rather than SSC as a standalone \emph{generation} task, and do not support \emph{controllable} evaluation over which stylistic dimensions should be described. This makes it difficult to compare SSC models consistently or evaluate whether models can selectively attend to requested style factors.

\subsection{Speaking-Style Captioning Models}
Advances in audio-language models enable speech-conditioned text generation \cite{cui2025voxeval}, typically by coupling a speech encoder with a pretrained LLM through an adapter \cite{verdini2024connect}. SSC-relevant approaches fall into two families. (1) \emph{General-purpose} audio-language models trained via multitask or instruction tuning \cite{das2024speechverse,chu2023qwen,11447408} can produce style-related descriptions \cite{huang2025speechcaps}, likely due to broad task coverage (e.g., ASR, AC, AST, SER, SV) \cite{tang2023salmonn}. (2) \emph{Task-specialized} SSC models explicitly target style caption generation: StyleCap \cite{yamauchi2024stylecap} combines a self-supervised speech encoder \cite{chen2022wavlm} with an auto-regressive LM; factor-conditioned methods introduce intermediate attribute prediction \cite{ando2024factor}; emotion-focused SSC variants emphasize free-form generation and multimodal prompting \cite{xu2024secap,xu2024multimodal}; and joint frameworks connect SSC with prompt-based speech synthesis \cite{zhu2024unistyle}. Despite this progress, prior work rarely provides a standardized setting to test instruction-level controllability across multiple stylistic factors, motivating our benchmark design.

\section{Methodology}

We first formalize controllable SSC as an instruction-conditioned generation task (Section~\ref{sec:formulation}), then describe \textit{StyleInstructCaps-DB}, the multi-dimensional caption dataset underlying the benchmark (Section~\ref{sec:dataset}). The evaluation tasks follow in Section~\ref{sec:benchmark}.





\subsection{Problem Formulation}
\label{sec:formulation}
We define \textit{controllable SSC} as generating, for a given speech signal, a natural-language caption that describes only the speaking-style dimension requested by an explicit instruction (e.g., speaker traits, emotional expression, or delivery style). Unlike traditional SSC, which produces a single holistic description per utterance, controllable SSC requires models to selectively ground instruction-relevant attributes while avoiding unsupported or irrelevant ones.


Let $\mathcal{D}=\{(x_i,s_i,y_i)\}_{i=1}^{N}$ contain speech utterances $x_i$, free-form instructions $s_i$, and reference captions $y_i=(y_{i,1},\dots,y_{i,T_i})$. An audio-language model with parameters $\theta$ generates $\hat{y}\sim p_\theta(y\mid x,s)$. \textit{StyleInstructCaps-DB} covers six speaking-style dimensions (Section~\ref{sec:dataset}), and the benchmark adds unseen instructions (Section~\ref{sec:benchmark}).


\begin{figure*}[!th]
        \centering
        \scalebox{1.0}{
        \includegraphics[draft=false,width=\linewidth]{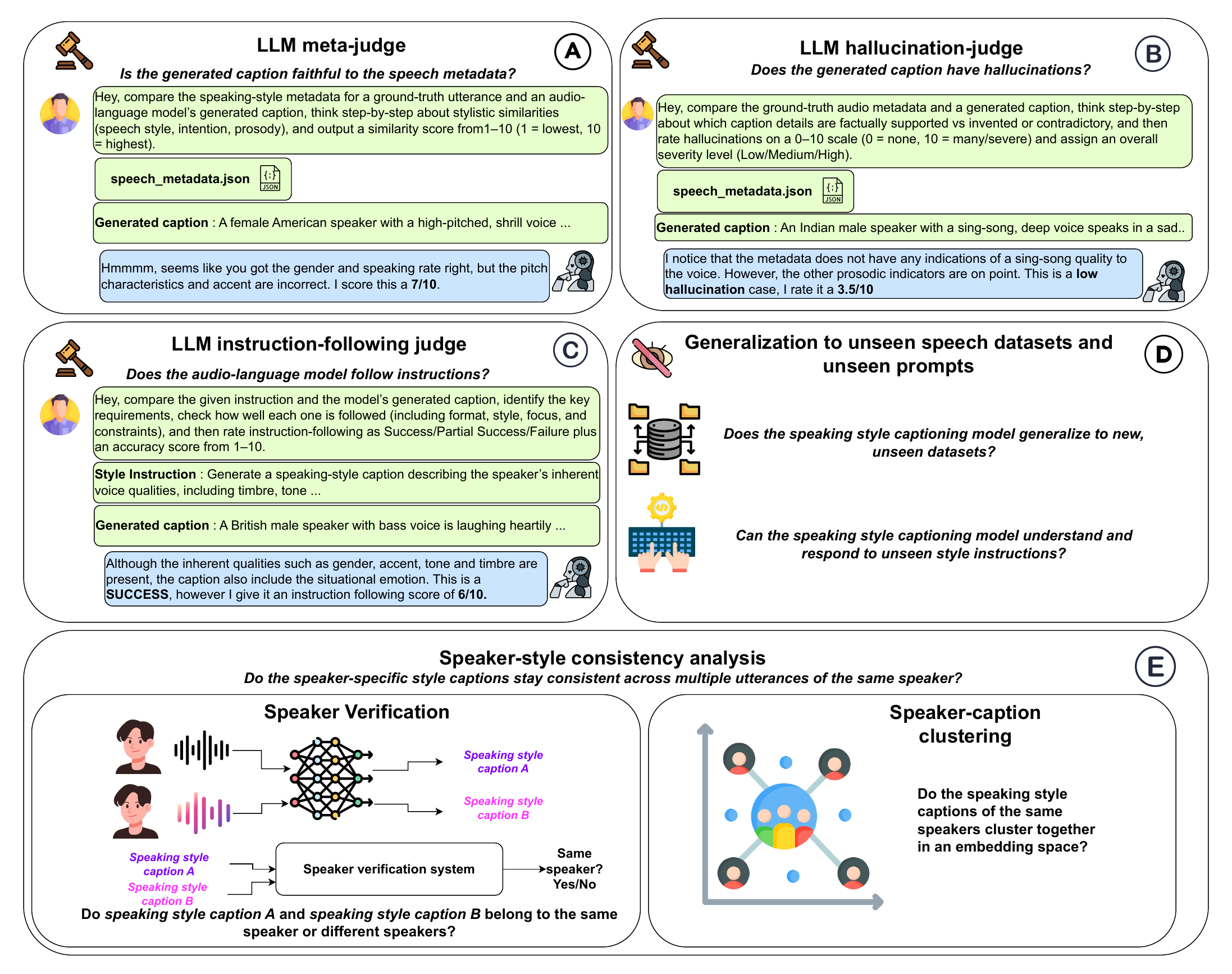}}
        \vspace{-10mm}
        \caption{Overview of \textit{StyleInstructCaps}, which evaluates models on metadata groundedness, hallucination, instruction-following, generalization to unseen prompts and speech data, and speaker-style consistency.}
        \label{fig:benchmark}
        \vspace{-5mm}
\end{figure*} 

\subsection{Dataset Description} 
\label{sec:dataset}
We introduce \textit{StyleInstructCaps-DB}, a multi-style speaking-style caption dataset built on the existing SSC corpus ParaSpeechCaps (PSC) \cite{diwan-etal-2025-scaling}. PSC aggregates multiple speech datasets and comprises two subsets: PSC-base, which includes Expresso \cite{nguyen2023expresso}, EARS \cite{richter2024ears}, and VoxCeleb \cite{nagrani2017voxceleb} and is human-annotated, and PSC-scaled, which consists of the Emilia dataset \cite{he2024emilia} and is automatically annotated. These annotations cover diverse intrinsic and situational speaking-style attributes across speakers and recording conditions. A brief overview of the PSC annotation procedure is provided in Appendix \ref{app:PSC_explain}.



While PSC provides a single caption for most utterances, \textit{StyleInstructCaps-DB} introduces six complementary stylistic captions per utterance: (1) speaker idiosyncratic style, (2) situational or contextual delivery style, (3) expressive or emotional style, (4) linguistic-pragmatic style, (5) perceptual listener-centric style, and (6) an overall style summary. These caption types jointly span representative dimensions of speaking-style used in speech understanding and generation tasks.


Captions are generated using an LLM conditioned on the rich per-utterance metadata provided by PSC; an example metadata record is shown in the Appendix \ref{app:dataset_details} Figure~\ref{fig:ears-example-metadata-p047}. \textit{StyleInstructCaps-DB} is organized into train, validation, and evaluation splits. We use Mistral-Small-24B-Instruct \cite{Mistral_Small_3} to generate captions for the train and validation splits, and Llama-3.1-8B-Instruct \cite{grattafiori2024llama3herdmodels} for the evaluation split, ensuring that benchmark evaluation does not benefit from exposure to captions generated by the same language model. Appendix \ref{app:dataset_details} Figure~\ref{prompt:caption_generation} presents the captioning prompt, including explicit definitions of the six speaking-style dimensions, while Appendix  \ref{app:dataset_details} Table~\ref{tab:style_instructions_captions}  reports instruction-caption pairs for the same utterance shown in Appendix \ref{app:dataset_details} Figure~\ref{fig:ears-example-metadata-p047}. 

\section{StyleInstructCaps Benchmark}
\label{sec:benchmark}
We introduce \textit{StyleInstructCaps}, a benchmark built on \textit{StyleInstructCaps-DB} for evaluating SSC models. The benchmark assesses captions along five evaluation dimensions (Figure~\ref{fig:benchmark}): metadata groundedness, hallucination, instruction-following, generalization to unseen instructions and speech data, and speaker-style consistency. The following subsections describe the setup and evaluation for each dimension.

\subsection{Metadata Groundedness}
\label{sec:metadata_ground}
In the first evaluation task (Figure~\ref{fig:benchmark}, Part~A), we assess \textbf{metadata groundedness}, the extent to which a generated caption is supported by and consistent with the speech metadata. We sample utterances from the \textit{StyleInstructCaps-DB} evaluation split with an emphasis on metadata richness, first selecting those with non-empty situational tags and then completing the set from the remaining pool. To ensure diversity, we cap each speaker at 10 utterances and limit any single accent to 25\% of the set. We refer to this subset as \textit{StyleInstructCaps-MetaSet} (distribution plots for accent, emotion, signal-to-noise ratio (SNR), and speech duration in Appendix~\ref{app:dataset_details}, Figure~\ref{fig:distribution_plots}). Using the overall style summary instruction, an LLM judge panel compares the generated caption against PSC ground-truth metadata on speaker gender, accent, pitch, speaking rate, environment or noise level, and emotion or expression, with particular emphasis on prosodic alignment. The judge assigns a 1--10 similarity score with a brief justification, and we report the mean across utterances as the meta-score (prompt in Appendix~\ref{app:prompts_used}, Figure~\ref{prompt:metajudge}).

\vspace{-2mm}

\subsection{Hallucination}
\label{sec:hallucination}

In the second evaluation task (Figure~\ref{fig:benchmark}, Part~B), we assess hallucination in generated speaking-style captions. We define hallucination as the inclusion of speaking-style attributes that are unsupported by or contradictory to the input speech. Evaluation is conducted on \textit{StyleInstructCaps-MetaSet}.

The LLM judge compares factual claims in the generated caption against the corresponding speech metadata, focusing on attributes such as speaker identity, emotion, accent, and acoustic characteristics. Descriptive or prosodic interpretations that do not contradict the metadata are treated leniently. The judge outputs a hallucination score on a 0 - 10 scale, together with a categorical severity label (Low, Medium, or High), reflecting both the frequency and severity of hallucinated attributes. The prompt is provided in Appendix \ref{app:prompts_used} Figure~\ref{fig:prompt-hallucination-eval}.

\subsection{Instruction-Following}
\label{sec:instruction_following}


We evaluate the controllability of SSC models through an explicit instruction-following task (See Figure \ref{fig:benchmark} Part C). Using the \textit{StyleInstructCaps-MetaSet} set, we assess whether models generate captions that adhere to user-specified style instructions. This evaluation probes instruction-following ability across the six speaking-style dimensions defined in \textit{StyleInstructCaps-DB}, testing whether models can selectively attend to and describe the intended stylistic attributes. The LLM-based judge evaluates each generated caption for compliance with the given instruction, considering content relevance, stylistic focus, and constraint satisfaction. The judge outputs a categorical instruction-following outcome (Success or Failure) and a scalar score on a 1--10 scale, accompanied by a brief justification. We report the mean of this score as the \emph{IFA score} (instruction-following accuracy, 1--10) and the percentage of captions judged a Success as the \emph{IFA rate} (\%). The complete evaluation prompts are provided in Appendix \ref{app:prompts_used} Figure ~\ref{fig:prompt-instruction-following-eval-a} and~\ref{fig:prompt-instruction-following-eval-b}.

\subsection{Generalization to Unseen Instructions}

Beyond the six speaking-style dimensions in \textit{StyleInstructCaps-MetaSet}, we assess generalization to instructions that differ \emph{both in surface form and output requirements} from those seen during training. Training instructions are primarily narrative directives, whereas the unseen set is deliberately heterogeneous and includes short, deployment-oriented controls: slot-/tag-like schemas (e.g., YAML or Python-dict templates), strict JSON outputs, hard lexical and length constraints, numeric rating scales, and persona-conditioned prompts (e.g., call-center evaluator). This tests whether instruction tuning yields genuine instruction-following ability rather than overfitting to the verbose narrative style of the training distribution. We pair two utterances from each of 16 accent-diverse VCTK speakers~\cite{yamagishi2019cstr} with all 36 unseen instructions (Appendix~\ref{app:prompts_used}, Figure~\ref{fig:unseen-instruction-prompts}). Each of the resulting 1{,}152 outputs per system is scored with the instruction-following protocol and Qwen3-32B judge of Section~\ref{sec:instruction_following}, and we report the mean 1--10 score per system (Table~\ref{tab:ood_prompts}).

\subsection{Generalization to Unseen Speech Dataset}
To evaluate SSC generalization to unseen speech datasets, we construct an evaluation set using the GigaSpeech corpus \cite{chen2021gigaspeech} with corresponding captions from Speechcraft \cite{jin2024speechcraft}. We sample 600 speech segments (300 from YouTube and 300 from podcasts), ensuring each segment originates from a distinct episode and is at least 3 seconds long. Each utterance is paired with a randomly sampled overall style summary speaking-style instruction, enabling evaluation under realistic acoustic conditions and diverse recording environments. Following the protocols in Sections \ref{sec:metadata_ground} and \ref{sec:hallucination}, we assess metadata groundedness and hallucination with respect to the SpeechCraft captions. We note that Speechcraft provides limited metadata coverage compared to PSC.


\begin{table*}[t]
\centering

\small
\caption{
Metadata groundedness, hallucination severity, and instruction-following performance evaluated on \textit{StyleInstructCaps-MetaSet}.
Blue shading indicates relative performance for scalar metrics. Metadata groundedness and hallucination are averaged over a panel of three LLM judges (Qwen3-32B, Gemma-4-31B-it, GPT-OSS-20B) with three runs per judge (9 runs per model); instruction-following is scored with a single Qwen3-32B run.
}
\small
\setlength{\tabcolsep}{5pt} 
\renewcommand{\arraystretch}{1.05}

\scalebox{0.8}{
\begin{tabularx}{\textwidth}{
>{\raggedright\arraybackslash}p{0.245\textwidth}
>{\centering\arraybackslash}p{0.095\textwidth}
>{\centering\arraybackslash}p{0.095\textwidth}
>{\centering\arraybackslash}p{0.060\textwidth}
>{\centering\arraybackslash}p{0.060\textwidth}
>{\centering\arraybackslash}p{0.060\textwidth}
>{\centering\arraybackslash}p{0.095\textwidth}
>{\centering\arraybackslash}p{0.095\textwidth}
}
\toprule
\multirow{2}{*}{\textbf{Model name}} &
\multicolumn{2}{c}{\makecell{\textbf{LLM metadata}\\\textbf{groundedness}}} &
\multicolumn{3}{c}{\makecell{\textbf{LLM hallucination}\\\textbf{severity (\%)}}} &
\multicolumn{2}{c}{\makecell{\textbf{Instruction}\\\textbf{following}}} \\
\cmidrule(lr){2-3} \cmidrule(lr){4-6} \cmidrule(lr){7-8}
& \textbf{\mbox{Meta score$\uparrow$}}
& \textbf{\mbox{Hall.\,$\downarrow$}}
& \textbf{Low}
& \textbf{Mid}
& \textbf{High}
& \textbf{\mbox{IFA score$\uparrow$}}
& \textbf{\mbox{IFA rate(\%)$\uparrow$}} \\
\midrule
Oracle Caption (PSC) & 9.48
&
0.03 &
99.5 & 0.5 & 0 & - 
& - 
\\

\makecell[l]{Oracle caption (\textit{SIC-DB})} &9.86
 &
0.22 &
99.8 & 0.2 & 0 &8.48  & 71.3
\\

\midrule

Audio-Flamingo-3 &
\cellcolor{blue0}3.88 &
\cellcolor{blue1}6.43 &
15.4 & 20.7 & 63.8 &
\cellcolor{blue2}8.21 &
\cellcolor{blue2}62.4 \\

Voxtral-Mini-3B &
\cellcolor{blue1}4.99 &
\cellcolor{blue3}4.29 &
48.6 & 35.0 & 16.4 &
\cellcolor{blue2}8.15 &
\cellcolor{blue3}68.0 \\

Voxtral-Small-24B &
\cellcolor{blue1}4.38 &
\cellcolor{blue4}1.99 &
73.5 & 16.8 & 9.68 &
\cellcolor{blue2}7.51 &
\cellcolor{blue3}67.8 \\

Qwen2-Audio-7B-Instruct &
\cellcolor{blue0}3.62 &
\cellcolor{blue0}7.00 &
19.1 & 27.9 & 53.0 &
\cellcolor{blue1}6.58 &
\cellcolor{blue1}53.6 \\

MERaLiON-AudioLLM &
\cellcolor{blue0}3.62 &
\cellcolor{blue4}1.26 &
81.8 & 10.6 & 7.60 &
\cellcolor{blue1}6.11 &
\cellcolor{blue0}27.7 \\

MOSS-Audio &
\cellcolor{blue3}6.32 &
\cellcolor{blue1}5.50 &
21.8 &
45.4 &
32.7 &
\cellcolor{blue3}8.82 &
\cellcolor{blue3}82.8 \\

Qwen3-Omni &
\cellcolor{blue2}5.75 &
\cellcolor{blue3}3.86 &
51.2 &
30.6 &
18.2 &
\cellcolor{blue1}7.99 &
\cellcolor{blue0}74.0 \\

Audio-Flamingo-next &
\cellcolor{blue2}5.09 &
\cellcolor{blue1}5.54 &
19.35 &
38.94 &
41.7 &
\cellcolor{blue1}7.59 &
\cellcolor{blue0}48.5 \\
SALMONN &
\cellcolor{blue1}4.68 &
\cellcolor{blue2}4.66 &
42.1 & 28.3 & 29.5 &
\cellcolor{blue2}7.50 &
\cellcolor{blue2}58.5 \\

\midrule
SALMONN (PSC) &
\cellcolor{blue4}7.01 &
\cellcolor{blue2}5.37 &
21.2 & 46.3 & 32.5 &
\cellcolor{blue2}7.98 &
\cellcolor{blue1}52.9 \\

\makecell[l]{SALMONN (\textit{SIC-DB}) - ST} &
\cellcolor{blue4}7.30 &
\cellcolor{blue3}4.33 &
35.2 & 37.3 & 27.4 &
\cellcolor{blue3}8.37 &
\cellcolor{blue3}65.6 \\

\makecell[l]{SALMONN (\textit{SIC-DB}) - MT} &
\cellcolor{blue4}7.35 &
\cellcolor{blue3}4.04 &
41.3 & 38.2 & 20.5 &
\cellcolor{blue3}8.62 &
\cellcolor{blue4}75.0 \\

\bottomrule
\end{tabularx}
}
\par\vspace{3pt}
{\footnotesize \textit{Closed-source reference} --- OpenAI-GPT-Audio: Meta score$\uparrow$ 5.70, Hall.$\downarrow$ 5.14.\par}
\label{tab:LLM_judge_3_evals}
\vspace{-3mm}

\end{table*}


\definecolor{blue0}{RGB}{255,255,255}   
\definecolor{blue1}{RGB}{240,247,255}
\definecolor{blue2}{RGB}{228,240,252}
\definecolor{blue3}{RGB}{214,232,248}
\definecolor{blue4}{RGB}{198,224,244}

\begin{table*}[!t]
\centering
\caption{Speaker-style consistency evaluation using clustering and verification metrics on the ID-balanced and UD-VCTK sets. Darker shading indicates better performance per column. Higher is better for ARI, NMI, Purity, InterSep, and ROC-AUC; lower is better for EER.}
\label{tab:speaker_style_consistency}
\resizebox{\textwidth}{!}{%
\begin{tabular}{lcccccc|cccccc}
\toprule
\multirow{2}{*}{\textbf{Model}} &
\multicolumn{6}{c|}{\textbf{ID-balanced}} &
\multicolumn{6}{c}{\textbf{UD-VCTK}} \\
\cmidrule(lr){2-7} \cmidrule(lr){8-13}
& ARI$\uparrow$ & NMI$\uparrow$ & Purity$\uparrow$ & InterSep$\uparrow$ & EER$\downarrow$ & AUC$\uparrow$
& ARI$\uparrow$ & NMI$\uparrow$ & Purity$\uparrow$ & InterSep$\uparrow$ & EER$\downarrow$ & AUC$\uparrow$ \\
\midrule

Audio-Flamingo-3
& \pink{0}{0.009} & \pink{0}{0.046} & \pink{0}{0.100} & \pink{0}{0.046} & \pink{0}{0.476} & \pink{0}{0.532}
& \pink{0}{0.011} & \pink{0}{0.048} & \pink{0}{0.121} & \pink{0}{0.057} & \pink{0}{0.463} & \pink{0}{0.556} \\

Voxtral-Mini-3B
& \pink{0}{0.033} & \pink{0}{0.122} & \pink{0}{0.152} & \pink{0}{0.133} & \pink{1}{0.399} & \pink{1}{0.643}
& \pink{0}{0.003} & \pink{0}{0.028} & \pink{0}{0.103} & \pink{0}{0.063} & \pink{0}{0.486} & \pink{0}{0.519} \\

Voxtral-Small-24B
& \pink{1}{0.067} & \pink{1}{0.182} & \pink{1}{0.186} & \pink{1}{0.157} & \pink{1}{0.363} & \pink{1}{0.674}
& \pink{0}{0.011} & \pink{0}{0.052} & \pink{0}{0.119} & \pink{0}{0.076} & \pink{0}{0.480} & \pink{0}{0.532} \\

Qwen2-Audio-7B-Instruct
& \pink{0}{-0.002} & \pink{0}{0.010} & \pink{0}{0.072} & \pink{0}{0.026} & \pink{0}{0.498} & \pink{0}{0.499}
& \pink{0}{-0.002} & \pink{0}{0.007} & \pink{0}{0.083} & \pink{0}{0.025} & \pink{0}{0.496} & \pink{0}{0.498} \\

MERaLiON-AudioLLM
& \pink{1}{0.081} & \pink{2}{0.276} & \pink{1}{0.213} & \pink{3}{0.233} & \pink{1}{0.428} & \pink{1}{0.604}
& \pink{1}{0.039} & \pink{1}{0.146} & \pink{1}{0.167} & \pink{1}{0.169} & \pink{0}{0.476} & \pink{0}{0.534} \\
MOSS-Audio
& \pink{3}{0.191} & \pink{3}{0.381} & \pink{3}{0.347} & \pink{3}{0.221} & \pink{3}{0.282} & \pink{3}{0.792}
& \pink{2}{0.076} & \pink{2}{0.228} & \pink{2}{0.186} & \pink{2}{0.175} & \pink{3}{0.319} & \pink{3}{0.755} \\

Qwen3-Omni
& \pink{2}{0.152} & \pink{2}{0.297} & \pink{2}{0.243} & \pink{2}{0.204} & \pink{1}{0.376} & \pink{1}{0.630}
& \pink{2}{0.069} & \pink{2}{0.216} & \pink{2}{0.184} & \pink{1}{0.165} & \pink{3}{0.311} & \pink{2}{0.734} \\

Audio-Flamingo-Next
& \pink{2}{0.113} & \pink{2}{0.318} & \pink{2}{0.242} & \pink{3}{0.227} & \pink{2}{0.348} & \pink{2}{0.717}
& \pink{2}{0.086} & \pink{3}{0.264} & \pink{2}{0.187} & \pink{3}{0.212} & \pink{1}{0.385} & \pink{1}{0.673} \\

SALMONN
& \pink{4}{0.351} & \pink{4}{0.598} & \pink{4}{0.511} & \pink{4}{0.290} & \pink{4}{0.225} & \pink{4}{0.854}
& \pink{3}{0.145} & \pink{3}{0.324} & \pink{3}{0.303} & \pink{2}{0.205} & \pink{3}{0.320} & \pink{3}{0.749} \\

\midrule

\makecell[l]{SALMONN (PSC)}
& \pink{4}{0.297} & \pink{4}{0.541} & \pink{4}{0.491} & \pink{4}{0.271} & \pink{4}{0.231} & \pink{4}{0.861}
& \pink{4}{0.157} & \pink{4}{0.340} & \pink{4}{0.312} & \pink{3}{0.209} & \pink{3}{0.320} & \pink{2}{0.711} \\

\makecell[l]{SALMONN(\textit{SIC-DB}) - ST}
& \pink{4}{0.363} & \pink{4}{0.577} & \pink{4}{0.508} & \pink{4}{0.284} & \pink{4}{0.214} & \pink{4}{0.876}
& \pink{3}{0.145} & \pink{3}{0.324} & \pink{3}{0.303} & \pink{2}{0.205} & \pink{3}{0.320} & \pink{3}{0.749} \\

\makecell[l]{SALMONN(\textit{SIC-DB}) - MT}
& \pink{4}{0.357} & \pink{4}{0.590} & \pink{4}{0.517} & \pink{4}{0.279} & \pink{4}{0.212} & \pink{4}{0.876}
& \pink{4}{0.156} & \pink{4}{0.341} & \pink{3}{0.298} & \pink{4}{0.222} & \pink{4}{0.282} & \pink{4}{0.795} \\

\bottomrule
\end{tabular}%
}
\end{table*}

\subsection{Speaker-Style Consistency}

Speaker-specific style captions describe inherent vocal traits and enable text-prompted speech synthesis via natural language control rather than explicit speaker IDs \cite{wang2025capspeech}.


To evaluate speaker-level consistency, we design two complementary evaluation schemes (See Figure \ref{fig:benchmark} Part E). \textit{Speaker-caption verification} measures pairwise consistency by testing whether captions generated from different utterances preserve invariant speaker-specific attributes. \textit{Speaker-caption clustering analysis} evaluates the global structure of speaker-related information in the caption embedding space, examining whether captions from the same speaker form coherent and well-separated clusters.

Experiments are conducted on two datasets. The first is an in-domain balanced set (ID-Balanced) from the \textit{StyleInstructCaps-DB} evaluation split, consisting of 400 utterances from speakers with at least 20 utterances each, with balanced gender and accent coverage (See Appendix \ref{app:dataset_details} Figure \ref{fig:three-side-by-side} for distribution plots). The second is an unseen-speaker set (UD-VCTK) derived from VCTK, containing 320 utterances from 16 speakers spanning diverse accents. In both settings, each utterance is paired with the same 10 speaker-specific instructions applied to assess robustness to unseen speakers, accents, and recording conditions.

\paragraph{Speaker-Caption Verification.}
We formulate speaker-caption verification as a binary verification task in the text embedding space, analogous to speaker verification in the speech domain \cite{5545402}. Positive trials pair captions from different utterances of the same speaker, while negative trials pair captions from different speakers. Captions are encoded using the pretrained E5-base-v2 text encoder \cite{wang2024textembeddingsweaklysupervisedcontrastive}, followed by attention-masked average pooling and L2 normalization. Speaker consistency is measured using cosine similarity between caption embeddings, with performance evaluated by Equal Error Rate (EER) and ROC-AUC.

\paragraph{Speaker-Caption Clustering Analysis.}
To assess global speaker organization, caption embeddings are clustered using k-means, with the number of clusters set equal to the number of speakers. Clustering quality is evaluated using adjusted rand index (ARI), normalized mutual information (NMI), and cluster purity. Inter-cluster separation is additionally reported to quantify global separation between speakers.

\subsection{Are LLM judges reliable for the SSC task evaluation?}
For metadata groundedness, hallucination, instruction-following, and generalization, we adopt the LLM-as-a-judge paradigm \cite{zheng2023judging}, which prior work has shown to align better with human preferences than automatic metrics such as ROUGE \cite{lin2004rouge}, CIDEr \cite{vedantam2015cider}, BERTScore \cite{zhang2019bertscore}, and n-gram based measurements \cite{mondshine2025beyond}. We validate this choice with three experiments.

\textbf{Adversarial gaming check.} Following the LLM judge reliability protocol of EmoSura~\cite{jing2026emosura}, we introduce controlled poisoning perturbations into the oracle captions. The resulting score shifts show that the judges detect such degradation in caption quality (Appendix~\ref{app:LLM_judge_adversarial}).

\textbf{Panel of LLM judges.} To ensure that the reported scores and system rankings are not an artifact of a single evaluator, we use the panel of three judges described in Section~\ref{sec:experiment_setup}. All judge pairs rank the systems consistently (Spearman $\rho \geq 0.864$), so the metadata-groundedness and hallucination scores in Tables~\ref{tab:LLM_judge_3_evals} and~\ref{tab:ood_eval} are averaged over the full panel (Appendix~\ref{app:judge_panel}).

\textbf{Human agreement with the judge.} In a pairwise preference study, thirteen listeners made forced choices between captions of different SSC systems over 100 caption pairs spanning all 14 systems in Table~\ref{tab:LLM_judge_3_evals}, while hearing only the raw audio. The human majority vote agrees with the judge panel on 85.9\% of pairs (Cohen's $\kappa = 0.717$), rising from 73.3\% on near-ties to 93.3\% on the largest score gaps, and a Bradley--Terry ranking fit to the human votes recovers the judge system ordering at Spearman $\rho = 0.895$. The protocol and full results are given in Appendix~\ref{app:human_llm_correlation}.

\section{Experiment Setup}
\label{sec:experiment_setup}

We use a panel of three open-weight LLM judges from different model families: Qwen3-32B \cite{yang2025qwen3} (thinking), Gemma-4-31B-it (thinking) \cite{team2026gemma}, and GPT-OSS-20B (reasoning) \cite{openai2025gptoss120bgptoss20bmodel}. Open-weight judges enable reproducible, open evaluation compared to closed-source alternatives. Metadata groundedness and hallucination are averaged over this panel, with three runs per judge on \textit{StyleInstructCaps-MetaSet} and two runs per judge on the out-of-domain evaluation sets (GigaSpeech, VCTK). Instruction-following is scored with a single Qwen3-32B run because of its high computational cost of 2{,}604 judgments per audio-language model, judge, and run (Appendix~\ref{app:judge_panel}).

We evaluate nine open-source audio-language models: Audio-Flamingo-3, Voxtral-Mini-3B, Voxtral-Small-24B, Qwen2-Audio-7B-Instruct, MERaLiON-AudioLLM, Audio-Flamingo-Next, MOSS-Audio, Qwen3-omni and SALMONN (specifications in Appendix \ref{app:model_training}, Table~\ref{table:models}). As a closed-source reference, we additionally evaluate OpenAI-GPT-Audio; we report only its core groundedness metrics, as the fine-grained analyses require large-scale inference with a paid API. We further instruction-tune SALMONN under three configurations: (1) using the original ParaSpeechCaps captions, SALMONN (PSC); (2) using the overall style summary captions from \textit{StyleInstructCaps-DB} in a single-task setup, SALMONN (\textit{SIC-DB})-ST; and (3) using the multi-caption set from SIC-DB with multiple instruction-caption pairs per utterance in a multi-task setup, SALMONN (\textit{SIC-DB})-MT. All variants are trained with supervised fine-tuning (SFT), minimizing the token-level negative log-likelihood over their respective training set $\mathcal{D}$ (Section~\ref{sec:formulation}):
\[
\begin{aligned}
\mathcal{L}_{\text{SFT}}(\theta)&=\mathbb{E}_{(x,s,y)\sim\mathcal{D}}\\
&\qquad\left[\sum_{t=1}^{T}-\log p_\theta\!\left(y_t\mid x,s,y_{<t}\right)\right].
\end{aligned}
\]
 Additional training and inference details are in Appendix \ref{app:model_training}. Data pre-processing follows the ParaSpeechCaps setup \cite{diwan-etal-2025-scaling}, and split statistics are in Appendix \ref{app:dataset_details}, Table \ref{tab:data_splits}. We further include an analysis of lexical
diversity of the generated captions in the Appendix
\ref{app:lexical-diversity}, reporting HD-D, Self-BLEU, Distinct-2,
Distinct-3, and caption length statistics (mean
and standard deviation) - to measure the stylistic
richness of the speaking style captions.



\definecolor{g1}{RGB}{245, 252, 248}   
\definecolor{g2}{RGB}{232, 246, 238}
\definecolor{g3}{RGB}{216, 240, 226}
\definecolor{g4}{RGB}{196, 232, 212}
\definecolor{g5}{RGB}{176, 224, 198}

\newcommand{\g}[2]{\cellcolor{g#1}{#2}}




\section{Results and Discussion}

Table \ref{tab:LLM_judge_3_evals} reports the results for the first three evaluation tasks -- metadata groundedness, hallucination severity and instruction-following, all evaluated by LLM judges. Broadly, we observe that \textbf{general-purpose SOTA large audio-language models tend to struggle with the SSC task}, even though they have been trained in a large-scale setting with iterative training strategies. Metadata groundedness and hallucination severity are largely independent across models (Spearman $\rho \approx 0$): some weakly grounded models hallucinate rarely (e.g., MERaLiON-AudioLLM, Voxtral-Small-24B) while others hallucinate heavily (e.g., Qwen2-Audio, Audio-Flamingo-3), so a low hallucination score alone does not indicate faithful style description. SSC-specific training yields the largest gains in groundedness: all three SSC-trained SALMONN variants exceed the best general-purpose model, MOSS-Audio.  Appendix \ref{app:LLM_judgement}  Figure~\ref{fig:judge_record_p019_pretty}  provides an example of judgment for the metadata groundedness evaluation. Pairwise significance tests over utterances are reported in  (Appendix~\ref{app:significance}).

Interestingly, models trained with SSC-relevant audio encoders and tasks outperform those exposed only to ASR, which captures \textit{what} is said but not \textit{how} it is said. For instance, Audio-Flamingo-3 achieves state-of-the-art results on MMAU~\cite{sakshi2025mmau} and MMAR~\cite{NEURIPS2025_610a7d65} but shows a low metascore and high hallucination on SSC, while SALMONN, despite a smaller backbone, benefits from its BEATs~\cite{pmlr-v202-chen23ag} + Whisper~\cite{radford2023robust} encoder and SSC-relevant training tasks (SQA, GR, SER).




Next, we examine instruction-following performance. As shown in Table~\ref{tab:LLM_judge_3_evals}, instruction-following and metadata groundedness are moderately correlated across models,\footnote{Spearman $\rho = 0.68$ over the nine general-purpose models ($\rho = 0.73$ over all twelve systems); between hallucination severity and instruction-following, $\rho = 0.27$. Given the small number of systems, we report these as descriptive statistics.} but the relation is asymmetric: well-grounded models also follow instructions well, whereas strong instruction-following does not guarantee grounding. Hallucination severity is largely unrelated to instruction-following. Performance is primarily driven by the diversity of tasks seen during instruction tuning, the capacity of the underlying LLM backbone, and hyperparameter choices that prevent overfitting to narrow instruction distributions. MOSS-Audio is the strongest general-purpose model on both axes, showing that the two abilities can co-occur. In contrast, Audio-Flamingo-3 (IFA score 8.21) and Voxtral-Mini-3B (8.15) achieve strong instruction-following despite weak faithfulness-oriented performance (meta scores of 3.88 and 4.99). \textbf{This dissociation indicates that generic instruction tuning suffices to produce stylistically well-formed captions but fails to ground them in speech-specific paralinguistic attributes}. SSC-specific training alone does not improve instruction-following (IFA rate of 52.9\% for SALMONN (PSC) vs.\ 58.5\% for SALMONN), whereas instruction-diverse training does: SALMONN (\textit{SIC-DB})-MT raises the IFA score from 7.50 to 8.62 and the IFA rate from 58.5\% to 75.0\%. 

On the unseen speech dataset (Table~\ref{tab:ood_eval}), SSC-specific training brings no further groundedness gain: all SALMONN variants score 6.25--6.30, on par with base SALMONN (6.28; not significant, Appendix~\ref{app:significance}), while MOSS-Audio scores highest (6.70). Pooled instruction-following scores also hide a large effect of output format. We label each of the 36 unseen instructions by required format (narrative, key-value slots, tags and lists, minimal imperatives, persona) and re-score every system with the same Qwen3-32B judge. Table~\ref{tab:ood_prompts} shows that format sensitivity is model-specific: Audio-Flamingo-3 and Qwen2-Audio lose much of their controllability on key-value slots (3.34 and 4.00 vs.\ 7.26 and 7.27 on narrative), the Voxtral models drop on minimal imperatives (4.85 and 3.46), and MOSS-Audio and Qwen3-Omni remain strong across all formats. All SALMONN models, including the untuned baseline, score higher on key-value and tag/list schemas than on narrative instructions, so this pattern stems from the backbone rather than SSC-specific training.

\begin{table}[t]
  \centering
  \caption{Evaluation on the unseen speech dataset. Metadata groundedness (1--10, $\uparrow$) and hallucination count (0--10, $\downarrow$) are averaged over the three-judge panel (Qwen3-32B, Gemma-4-31B-it, GPT-OSS-20B) with two runs per judge (6 runs per model) on the GigaSpeech-derived out-of-domain evaluation set.}
  \label{tab:ood_eval}
  \vspace{-0.6em}
  \resizebox{\linewidth}{!}{%
    \begin{tabular}{lcc}
\toprule
\textbf{Model} & Meta score$\uparrow$ & Hall.$\downarrow$ \\
\midrule
OpenAI-GPT-Audio
& \g{2}{4.98} & \g{1}{7.02} \\
\midrule
Audio-Flamingo-3
& \g{1}{3.85} & \g{4}{5.97} \\
Voxtral-Mini-3B
& \g{2}{4.57} & \g{1}{7.12} \\
Voxtral-Small-24B
& \g{1}{3.45} & \g{3}{6.45} \\
Qwen2-Audio-7B-Instruct
& \g{2}{4.62} & \g{2}{6.97} \\
MERaLiON-AudioLLM
& \g{1}{3.95} & \g{4}{5.72} \\
MOSS-Audio
& \g{5}{6.70} & \g{4}{5.90} \\
Qwen3-Omni
& \g{2}{4.40} & \g{5}{3.81} \\
Audio-flamingo-next
& \g{3}{5.81} & \g{2}{6.65} \\
SALMONN
& \g{4}{6.28} & \g{3}{6.29} \\
\midrule
SALMONN (PSC)
& \g{4}{6.28} & \g{3}{6.38} \\
SALMONN (\textit{SIC-DB}) - ST
& \g{4}{6.25} & \g{4}{5.93} \\
SALMONN (\textit{SIC-DB}) - MT
& \g{4}{6.30} & \g{3}{6.05} \\
\bottomrule
\end{tabular}%
  }
  \vspace{-4mm}
\end{table}

\begin{table}[t]
  \centering
  \caption{Instruction-following on unseen prompts, broken down by the output format each instruction requires. Each cell aggregates 32 clips over the instructions in that family, giving 128 to 416 judgements per system and 1{,}152 per system in total.}
  \label{tab:ood_prompts}
  \vspace{-0.6em}
  \resizebox{\linewidth}{!}{%
    \begin{tabular}{lccccc}
\toprule
\textbf{Model} & Narrative & Key-value & Tags/list & Min. imper. & Persona \\
\midrule
Audio-Flamingo-3
& 7.26 & 3.34 & 4.12 & 6.39 & 7.47  \\
Voxtral-Mini-3B
& 6.23 & 6.30 & 7.31 & 4.85 & 7.41  \\
Voxtral-Small-24B
& 5.43 & 7.68 & 7.30 & 3.46 & 5.82  \\
Qwen2-Audio-7B-Instruct
& 7.27 & 4.00 & 5.63 & 5.84 & 6.96  \\
MERaLiON-AudioLLM
& 4.50 & 6.91 & 5.70 & 5.15 & 5.79  \\
MOSS-Audio
& 8.00 & 8.57 & 8.34 & 8.39 & 8.33  \\
Qwen3-Omni
& 7.08 & 8.89 & 9.07 & 7.89 & 7.72  \\
Audio-flamingo-next
& 6.69 & 7.62 & 7.52 & 6.38 & 7.06  \\
SALMONN
& 6.30 & 7.86 & 7.45 & 6.90 & 7.67  \\
\midrule
SALMONN (PSC)
& 6.57 & 7.77 & 7.52 & 5.35 & 7.14  \\
SALMONN (\textit{SIC-DB}) - MT
& 6.69 & 7.85 & 8.28 & 6.67 & 7.68 \\
\bottomrule
\end{tabular}%
  }
  \vspace{-4mm}
\end{table}

\begin{figure}
  \centering
  \vspace{0.15em}
  \includegraphics[draft=false,width=\linewidth]{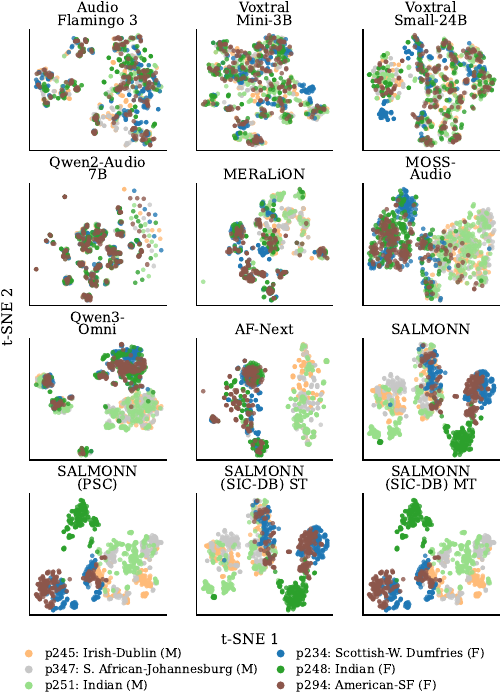}
  \vspace{-0.8cm}
  \captionof{figure}{t-SNE visualization of speaker-style caption embeddings on UD-VCTK (6 speaker subset)}
  \label{fig:ood_figure}
\end{figure}

Next, we turn to discussing speaker style consistency.
Table~\ref{tab:speaker_style_consistency} reveals clear and consistent trends in speaker-style consistency across both the ID-balanced set and UD-VCTK sets. \tealc{\textbf{SALMONN-based models yield the most speaker-consistent captions}, and this advantage is already present in the untuned baseline. All SALMONN variants outperform all other models on every metric on the ID-balanced set and on ARI, NMI, and Purity on UD-VCTK. On UD-VCTK, however, Qwen3-Omni and MOSS-Audio achieve lower EER than every variant except SALMONN (\textit{SIC-DB})-MT, which is best on EER (0.282) and AUC (0.795), and Audio-Flamingo-Next exceeds all variants but MT on InterSep. These results indicate that SALMONN-based captions encode} stable, speaker-specific paralinguistic attributes rather than surface-level descriptions. The t-SNE \cite{maaten2008visualizing} visualizations of speaker-caption embeddings for a subset of six speakers from the UD-VCTK set (Figure~\ref{fig:ood_figure}) \tealc{show more coherent and well-separated speaker clusters for SALMONN-based models than for the other models.} A full t-SNE analysis across all speakers is provided in Appendix \ref{app:spk_cluster} Figure~\ref{fig:tsne_allspks_all_models}.


\tealc{Relative to the untuned baseline, style-caption fine-tuning yields mixed, modest changes: SALMONN (PSC) improves clustering on UD-VCTK but not on the ID-balanced set, while the \textit{SIC-DB} variants improve verification (EER, AUC) on the ID-balanced set.} Single-task instruction tuning (ST) further improves consistency on the ID-balanced set but does not generalize to UD-VCTK, indicating over-specialization to a narrow instruction distribution. In contrast, \textbf{multi-task instruction tuning (MT) yields the most balanced and transferable performance}, with strong clustering structure and competitive verification scores across both datasets. Finally, the strong agreement between clustering and verification metrics indicates that these improvements reflect a well-structured caption embedding space. \tealc{\textbf{Overall, speaker consistency stems mainly from the SALMONN backbone; among fine-tuning strategies, instruction-diverse (MT) training transfers best across domains.}}


\definecolor{lightred}{RGB}{255, 235, 238}
\definecolor{lightblue}{RGB}{232, 244, 253}

\section{Conclusions}
\tealc{We introduced \textit{StyleInstructCaps}, a benchmark that formulates controllable SSC as an instruction-following audio-language task. Its dataset, \textit{StyleInstructCaps-DB}, extends ParaSpeechCaps with six complementary captions per utterance, covering speaker idiosyncratic, situational, expressive, linguistic-pragmatic, perceptual, and overall style. The benchmark evaluates models along five axes: metadata groundedness, hallucination, instruction-following, generalization to unseen instructions and speech datasets, and speaker-style consistency. For groundedness and hallucination, we designed an LLM-as-a-judge protocol that checks captions against per-utterance speech metadata. We validated this protocol with adversarial perturbation checks, consistent system rankings across a three-judge panel (Spearman $\rho \geq 0.864$), and a listening study in which the human majority vote agreed with the panel on 85.9\% of caption pairs. We measured speaker-style consistency directly in the caption embedding space, using speaker-caption verification and clustering to test whether captions of the same speaker preserve invariant vocal traits across utterances. Together, these components turn SSC from a loosely defined capability into a measurable research task.}
\newpage

\section*{Limitations}
This study is subject to several limitations inherent to large-scale speaking-style captioning (SSC). First, publicly available speech corpora with rich, fine-grained, per-utterance speaking-style annotations remain limited in both scale and diversity. As a result, SSC systems are often trained and evaluated on datasets with constrained stylistic coverage, which may not fully capture the variability of real-world expressive speech across speakers, domains, recording conditions, and cultural contexts.

Second, the proposed evaluation framework relies substantially on LLM-based judgment for assessing open-ended caption quality, including groundedness, hallucination, and instruction-following behavior. Although recent LLM judges exhibit strong reasoning and language understanding capabilities, their evaluations remain dependent on the specific judge model, prompting strategy, and the distribution of examples encountered during pretraining and instruction tuning. Consequently, evaluation outcomes may vary across alternative judge architectures or future generations of language models. To reduce this dependence, we average the open-ended metrics over a panel of three judges from different model families, whose system rankings agree closely (pairwise Spearman $\rho$ of 0.864--0.978; Appendix~\ref{app:judge_panel}).

Third, despite efforts to improve robustness through neutral prompt design, prompt paraphrasing, and multi-prompt evaluation, LLM-based assessments may still exhibit sensitivity to prompt wording, decoding configurations, or stylistic biases in generated text. While our adversarial calibration experiments demonstrate that the judge is sensitive to targeted perturbations, such analyses cannot fully eliminate all forms of evaluation instability or hidden biases.

Fourth, the overall pipeline introduces a potential closed-loop evaluation concern, since metadata-derived descriptions are used both to construct reference captions and to evaluate generated outputs through LLM-based scoring. Although we explicitly validate the sensitivity of the judge to adversarial manipulations and metadata inconsistencies, some degree of implicit alignment between the generation and evaluation processes may still remain. Future work should further investigate evaluation paradigms involving independent human annotation and audio-grounded verification.

Finally, the current benchmark primarily evaluates metadata-grounded captioning rather than fully audio-grounded speaking-style understanding. In many cases, the reference captions reflect structured metadata annotations associated with the utterance rather than exhaustively capturing all perceptually salient properties present in the speech signal itself. As a result, models may partially optimize toward reproducing metadata-aligned descriptions instead of developing deeper holistic representations of expressive speech. Bridging this gap between metadata supervision and truly audio-grounded style understanding remains an important direction for future SSC research.

\section*{Ethical Considerations}
In this work, we utilize open-source audio-language models to study the task of speaking-style captioning. We emphasize that the paralinguistic understanding capabilities of large language models remain an open research problem and are not yet sufficiently reliable for high-stakes use. Consequently, the outputs and evaluations presented in this paper should not be interpreted as substitutes for human judgment in scenarios that require factual accuracy, mental-state inference, or sensitive assessments of speaker intent or affect. Our results are intended to support research and benchmarking in controlled experimental settings, rather than deployment in applications where misinterpretation of paralinguistic cues could lead to harm.






\bibliography{custom}

\newpage

\appendix

\section{Existing Datasets and Benchmarks}
\label{app:existing_benchmarks}

\begin{table}[ht]
\centering
\small
\setlength{\tabcolsep}{1.5pt} 
\begin{tabular}{l@{}c@{}c@{}c}
\toprule
\textbf{Name} & \textbf{Size (hrs)}  \\
\midrule

ParlerTTS {\scriptsize \cite{lyth2024natural}}
    & 45k \\
PromptSpeech {\scriptsize \cite{guo2023prompttts}} & 37 (real)\\
Expresso {\scriptsize \cite{nguyen2023expresso}} & 47 \\
EARS {\scriptsize \cite{richter2024ears}}& 60 \\
TextrolSpeech {\scriptsize \cite{ji2024textrolspeech}} & 300\\
MEAD-TTS {\scriptsize \cite{guan2024mm}}& 36\\
SpeechCraft {\scriptsize \cite{jin2024speechcraft}}& 2.4k \\
LibriTTS-P {\scriptsize \cite{kawamura2024libritts}}& 600 \\
Dreamvoice {\scriptsize \cite{hai2024dreamvoice}}& 231\\
Capspeech {\scriptsize \cite{wang2025capspeech}} &33.6k\\
ParaSpeechCaps {\scriptsize \cite{diwan-etal-2025-scaling}} & 2.7k \\
\bottomrule
\end{tabular}
\caption{Summary of existing datasets relevant to SSC.}
\label{tab:dataset_summary}
\end{table}

\begin{table*}[!ht]
\centering
\resizebox{0.9\textwidth}{!}{%
\begin{tabular}{p{3cm} p{7cm} p{5cm}}
\toprule
\textbf{Benchmark} & \textbf{What Task It Benchmarks} & \textbf{Evaluation Metrics} \\
\midrule

CP-Bench \cite{wang-etal-2025-benchmarking-contextual} & Contextual Paralinguistic Reasoning: Evaluates the model's ability to reason over both verbal content and non-verbal cues like emotion and prosody to answer questions & Answer accuracy, human preference ratings \\

SD-Eval \cite{ao2024sd} & Spoken Dialogue Understanding: assesses understanding of dialogue that relies on paralinguistic cues (e.g., sarcasm, tone) for correct interpretations & Task accuracy, BLEU (for generation), MOS-style human ratings. \\

ParaS2SBench\cite{ICLR2026_734b2e77} & Paralinguistic-Aware Speech-to-Speech: Evaluates if generated speech responses maintain appropriate paralinguistic alignment (style/emotion) with the input & Content similarity, style similarity, MOS (Mean Opinion Score) \\

ParaLBench \cite{zhang2024paralbench} & General Computational Paralinguistics: A comprehensive test suite across multiple existing paralinguistic datasets to test generalizability & Accuracy, UAR (Unweighted Average Recall), F1 score, correlation metrics \\

BERSt \cite{tuttosi2026bersting} & Expressive Emotion Recognition: Specifically benchmarks emotion recognition in challenging conditions like shouted or highly expressive speech & SER (Speech Emotion Recognition) accuracy, UAR, WER (Word Error Rate - as a stress test) \\

EmoNet-Voice-Bench \cite{schuhmann2026emonetvoicefinegrainedexpertverifiedbenchmark} & Fine-grained Emotion Understanding: Tests understanding of nuanced emotions beyond basic categories (e.g., distinguishing "cold anger" from "hot anger") & Classification accuracy, macro-F1, confusion matrix analysis \\

StepEval-Audio-Paralinguistic \cite{wu2025step} & Multidimensional paralinguistic comprehension: Evaluates 11 distinct dimensions including gender, age, pitch, rhythm, and speaking rate. & Dimension-wise accuracy, human ratings \\
S2Cap \cite{ok2025s2cap} &
Singing Style Description: Focuses on describing singing voices using natural language, capturing attributes like mood, tempo, and timbre &
Classification accuracy on style factors, Captioning metrics: BLEU, METEOR, CIDEr \\ 

SpeechCaps \cite{huang2025speechcaps} &
Speaking-Style Captioning in multi-talker scenarios: Benchmarks the ability to generate captions that describe speaker-specific styles and prosody in multi-talker scenarios. & Instruction-following rate, overall accuracy, and conditional accuracy, with correctness and relevance judged via LLM-based semantic alignment (GPT-4o)\\

WildSpeech-Bench \cite{zhang2025wildspeech} & 
Natural / In-the-Wild Prosody: Benchmarks how models handle real-world paralinguistic variability, including stuttering, hesitations, interruptions, and extreme prosodic conditions (e.g., whispering vs. shouting). & 
UGPT-4o-based 1-10 scoring, human-alignment metrics (Pearson/R²/MSE), and UTMOS.
\\

Multi-Bench \cite{deng2025multi}& 
Multi-turn Emotional Dialogue: Tests whether models can maintain emotional consistency and empathy over a long conversation, unlike single-turn benchmarks. & 
Consistency score, empathy rating (Likert scale by human or LLM judge) \\

Salmon \cite{maimon2025salmon} & Acoustic consistency and semantic-acoustic alignment across attributes such as sentiment, speaker identity, background noise, and room acoustics using likelihood-based comparisons. & Uses pairwise likelihood preference scores (accuracy) for acoustic consistency and semantic-acoustic alignment tasks, with sWUGGY for spoken-content modeling and human agreement as an upper bound. \\ 
\bottomrule
\end{tabular}%
}
\caption{Benchmarks for Speech Paralinguistics Understanding}
\label{tab:paralinguistic_benchmarks}
\end{table*}

\section{Annotation reliability of PSC style metadata}
\label{app:PSC_explain}

The quality of PSC annotations is supported by multiple stages of annotation verification and filtering throughout the dataset construction pipeline. For PSC-Base, annotations were collected from qualified Amazon Mechanical Turk annotators who completed a dedicated qualification task designed around identifying speaking-style attributes, and annotations were aggregated across five annotators per speaker with agreement-based filtering applied before retaining tags.   For PSC-Scaled, automatic annotation combines complementary components including perceptual speaker similarity, expressivity filtering, semantic matching, and acoustic matching using an audio LLM-as-a-judge, rather than relying on a single classifier or heuristic.   Importantly, the paper further verifies scaled data quality through human evaluation, showing that PSC-Scaled achieves tag recall comparable to PSC-Base for both intrinsic and situational tags, supporting that the automatically scaled annotations remain aligned with human perceptual judgments. 

\FloatBarrier
\clearpage

\twocolumn[{%
\section{StyleInstructCaps-DB Construction details}
\label{app:dataset_details}

\vspace{-2mm}
\begin{minipage}{\textwidth}
\centering

\begin{tikzpicture}
\node[
  draw,
  rounded corners=2pt,
  line width=0.4pt,
  inner sep=6pt,
  text width=0.97\textwidth,
  align=left
] (meta) {%
\begin{minipage}{\linewidth}
\ttfamily\small
\setlength{\parindent}{0pt}
\setlength{\parskip}{0pt}
\setlength{\baselineskip}{0.93\baselineskip}

\{\\
\ \ \ \ "source": "ears",\\
\ \ \ \ "relative\_audio\_path": "p047/emo\_amazement\_freeform.wav",\\
\ \ \ \ "text\_description": [\\
\ \ \ \ \ \ \ \ " A female speaker's authoritative, nasal voice is crisp and high-pitched, conveyed in an American accent with a measured speed. Her speech is delivered in a slightly clean environment, yet she expresses awe with a loud and slightly elevated volume.",\\
\ \ \ \ \ \ \ \ " An American woman speaks with a high-pitched voice, expressing awe, in a slightly clean environment. Her speech is measured and deliberate."\\
\ \ \ \ ],\\
\ \ \ \ "transcription": " Oh, really? You were so good at that. I love to just sit here and listen to you play. It's so relaxing. I just can't believe how talented you are.",\\
\ \ \ \ "intrinsic\_tags": [\\
\ \ \ \ \ \ \ \ "authoritative",\\
\ \ \ \ \ \ \ \ "crisp",\\
\ \ \ \ \ \ \ \ "loud",\\
\ \ \ \ \ \ \ \ "american",\\
\ \ \ \ \ \ \ \ "nasal"\\
\ \ \ \ ],\\
\ \ \ \ "situational\_tags": [\\
\ \ \ \ \ \ \ \ "awed"\\
\ \ \ \ ],\\
\ \ \ \ "basic\_tags": [\\
\ \ \ \ \ \ \ \ "female",\\
\ \ \ \ \ \ \ \ "high-pitched",\\
\ \ \ \ \ \ \ \ "measured speed",\\
\ \ \ \ \ \ \ \ "slightly clean environment"\\
\ \ \ \ ],\\
\ \ \ \ "all\_tags": [\\
\ \ \ \ \ \ \ \ "american",\\
\ \ \ \ \ \ \ \ "authoritative",\\
\ \ \ \ \ \ \ \ "awed",\\
\ \ \ \ \ \ \ \ "crisp",\\
\ \ \ \ \ \ \ \ "female",\\
\ \ \ \ \ \ \ \ "high-pitched",\\
\ \ \ \ \ \ \ \ "loud",\\
\ \ \ \ \ \ \ \ "measured speed",\\
\ \ \ \ \ \ \ \ "nasal",\\
\ \ \ \ \ \ \ \ "slightly clean environment"\\
\ \ \ \ ],\\
\ \ \ \ "speakerid": "p047",\\
\ \ \ \ "name": "p047",\\
\ \ \ \ "duration": 9.8,\\
\ \ \ \ "gender": "female",\\
\ \ \ \ "accent": "american",\\
\ \ \ \ "pitch": "high-pitched",\\
\ \ \ \ "speaking\_rate": "measured speed",\\
\ \ \ \ "noise": "slightly clean environment",\\
\ \ \ \ "utterance\_pitch\_mean": 220.68133544921875,\\
\ \ \ \ "snr": 56.96498489379883,\\
\ \ \ \ "phonemes": " oʊ, ɹɪli? ju wɜ˞ soʊ ɡʊd æt ðæt. aɪ lʌv tu dʒʌst sɪt hiɹ ʌnd lɪsʌn tu ju pleɪ. ɪt'ɛs soʊ ɹɪlæksɪŋ. aɪ dʒʌst kæn'ti bɪliv haʊ tælʌntɪd ju ɑɹ.",\\
\ \ \ \ "audio\_path": "datasets/ears/p047/emo\_amazement\_freeform.wav",\\
\}\\
\end{minipage}
};
\end{tikzpicture}

\vspace{-1mm}
\captionof{figure}{Example metadata object from EARS used in our captioning/evaluation pipeline.}
\label{fig:ears-example-metadata-p047}

\end{minipage}
\vspace{2mm}
}]%
\newpage
\begin{figure*}
\centering
\begin{tikzpicture}
\node[
  draw,
  rounded corners=2pt,
  line width=0.4pt,
  inner sep=6pt,
  text width=0.87\textwidth,
  align=left
] (prompt) {%
\begin{minipage}{\linewidth}
\ttfamily\scriptsize

[ROLE \& GOAL]\\
You are an AI expert in speech prosody. Your primary task is to generate six highly descriptive, fine-grained speech captions from a given audio metadata object.\\[0.35em]

[CONTEXT: INPUT METADATA]\\
Analyze the following JSON object, which contains detailed metadata for a speech utterance.\\[0.35em]

JSON\\
\{metadata\}\\[0.6em]

[TASK \& INSTRUCTIONS]\\
Based on the metadata, generate six unique and descriptive captions. For each caption, combine multiple vocal characteristics using descriptive verbs and adverbs to precisely capture the essence of the speaking-style.\\
Use these general guidelines for each caption:\\[0.35em]

1. Think in Verbs and Adverbs: Instead of "sad," think "a voice that wavers and falls off slowly."\\[0.25em]
2. Combine at Least Three Elements: A good caption should ideally define a pitch contour, a pace, and a vocal quality.\\[0.25em]
3. Specify the Context: Adding a phrase like "as if explaining a complex topic" or "like a conspiratorial whisper" can give the model a powerful contextual anchor.\\[0.25em]
4. Use the Tags: Leverage the intrinsic\_tags, situational\_tags, basic\_tags, and all\_tags to inform your captions.\\[0.25em]
5. Always associate the nationality with the accent\\[0.25em]
6. Use the provided transcription to infer emotional and contextual elements but do not directly quote it in the captions.\\[0.35em]

Each caption should focus on a different aspect of the speech, as outlined below.\\[0.35em]

1. Speaker-Idiosyncratic Style:\\[0.25em]
Goal: Capture the speaker's core vocal identity.\\[0.15em]
Instruction: Generate a caption describing the speaker's inherent qualities. Combine the pitch , accent,  gender , and timbre-related intrinsic\_tags.\\[0.6em]

2. Situational/Contextual Style:\\[0.25em]
Goal: Describe the circumstances of the speech.\\[0.15em]
Instruction: Generate a caption that sets the scene. Combine the speaking\_rate (measured speed), the noise (noisy environment), and contextual intrinsic\_tags (authoritative). Frame it as a person speaking with purpose in a specific setting.\\[0.6em]

3. Expressive/Emotional Style:\\[0.25em]
Goal: Convey the inferred emotion or attitude.\\[0.15em]
Instruction: Generate a caption describing the emotional tone. Infer the attitude from the transcription, situational\_tags and all\_tags. Describe the prosody (pitch contour, rhythm) that would create this feeling.\\[0.6em]

4. Linguistic/Pragmatic Style:\\[0.25em]
Goal: Detail the speech's structural and emphasis patterns.\\[0.15em]
Instruction: Generate a caption focused on the delivery of the words themselves. Analyze the transcription to identify linguistic features. Describe the pausing and stress patterns.\\[0.6em]

5. Perceptual/Listener-Centric Style:\\[0.25em]
Goal: Describe the impression the voice leaves on a listener.\\[0.15em]
Instruction: Generate a caption from the listener's point of view. Use impressionistic intrinsic\_tags to describe the overall effect of the voice. How does it feel to hear this person speak?\\[0.6em]

6. Overall style summary:\\[0.25em]
Goal: Create a rich, narrative-style caption.\\[0.15em]
Instruction: Combine all elements in instrinsic-tags, situational\_tags, basic\_tags, all\_tags, gender, accent, pitch, speaking\_rate, noise into a creative, narrative-style caption that captures the essence of the speaking-style.\\[0.6em]

[OUTPUT FORMAT \& CONSTRAINTS]\\
Return your response as a single, valid JSON object. Use the descriptive keys provided in the example below. The entire output must be only the JSON object, with no additional text, explanations, or markdown formatting.\\[0.35em]

JSON\\[0.2em]
\{\\
\ \ "speaker\_idiosyncratic\_style": "\textless Generated caption for style 1\textgreater",\\
\ \ "situational\_contextual\_style": "\textless Generated caption for style 2\textgreater",\\
\ \ "expressive\_emotional\_style": "\textless Generated caption for style 3\textgreater",\\
\ \ "linguistic\_pragmatic\_style": "\textless Generated caption for style 4\textgreater",\\
\ \ "perceptual\_listener\_centric\_style": "\textless Generated caption for style 5\textgreater",\\
\ \ "overall\_style\_summary": "\textless Generated caption for style 6\textgreater"\\
\}\\
\end{minipage}
};
\end{tikzpicture}

\caption{Prompt template for generating six style-specific captions from an input metadata object.}
\label{prompt:caption_generation}
\end{figure*}
\begin{table*}[t]
\centering
\small
\setlength{\tabcolsep}{6pt}
\renewcommand{\arraystretch}{1.2}
\begin{tabular}{p{0.25\textwidth} p{0.30\textwidth} p{0.36\textwidth}}
\hline
\textbf{Instruction type} & \textbf{Style instruction} & \textbf{Generated caption} \\
\hline
Speaker idiosyncratic  style &
Describe the speaker's inherent vocal identity by combining their pitch, accent, gender, and timbre-related characteristics. &
A high-pitched, crisp, and nasal voice with an American accent, delivered by a female speaker, instantly conveys a sense of authority and confidence. \\
\hline
Situational contextual style &
Set the scene by combining the speaking rate, noise level, and contextual intrinsic tags to describe a person speaking with intention in a specific setting. &
Speaking with measured speed and authority, a female voice with an American accent cuts through a slightly clean environment, commanding attention as if addressing a gathering with purpose and conviction. \\
\hline
Expressive emotional style &
Infer the emotional tone from the transcription and situational tags, and describe the prosody that creates this feeling, capturing the speaker's attitude and sentiment. &
A voice filled with awe and admiration, conveyed through a rising pitch contour and deliberate pacing, expresses genuine astonishment, as if the speaker is witnessing something remarkable for the first time. \\
\hline
Linguistic pragmatic style &
Analyze the transcription to identify linguistic features, and describe the pausing and stress patterns that reveal the speaker's emphasis and intention. &
The speech unfolds with a steady, measured rhythm, punctuated by deliberate pauses that emphasize key phrases, as the speaker carefully chooses words to convey their thoughts, creating a sense of sincerity and authenticity. \\
\hline
Perceptual listener-centric style &
Describe the overall effect of the voice on the listener, using impressionistic intrinsic tags to capture the impression the voice leaves on those who hear it. &
Listening to this voice is like being drawn into a conversation with a charismatic and confident individual, as the speaker's authoritative tone and crisp delivery create a sense of trust and engagement. \\
\hline
Holistic creative synthesis &
Combine all elements to create a rich, narrative-style caption that captures the essence of the speaking-style, weaving together the speaker's inherent qualities, situational context, emotional tone, and linguistic features into a vivid and immersive portrait. &
Imagine a female speaker with a high-pitched, crisp, and nasal voice, speaking with measured speed and authority, her American accent shining through as she expresses awe and admiration in a slightly clean environment, conveying confidence and charisma with every word. \\
\hline
\end{tabular}
\caption{Style-specific instructions and example generated captions.}
\label{tab:style_instructions_captions}
\end{table*}

\begin{figure*}
    \centering
    \begin{minipage}{0.48\linewidth}
        \centering
        \includegraphics[draft=false,width=\linewidth]{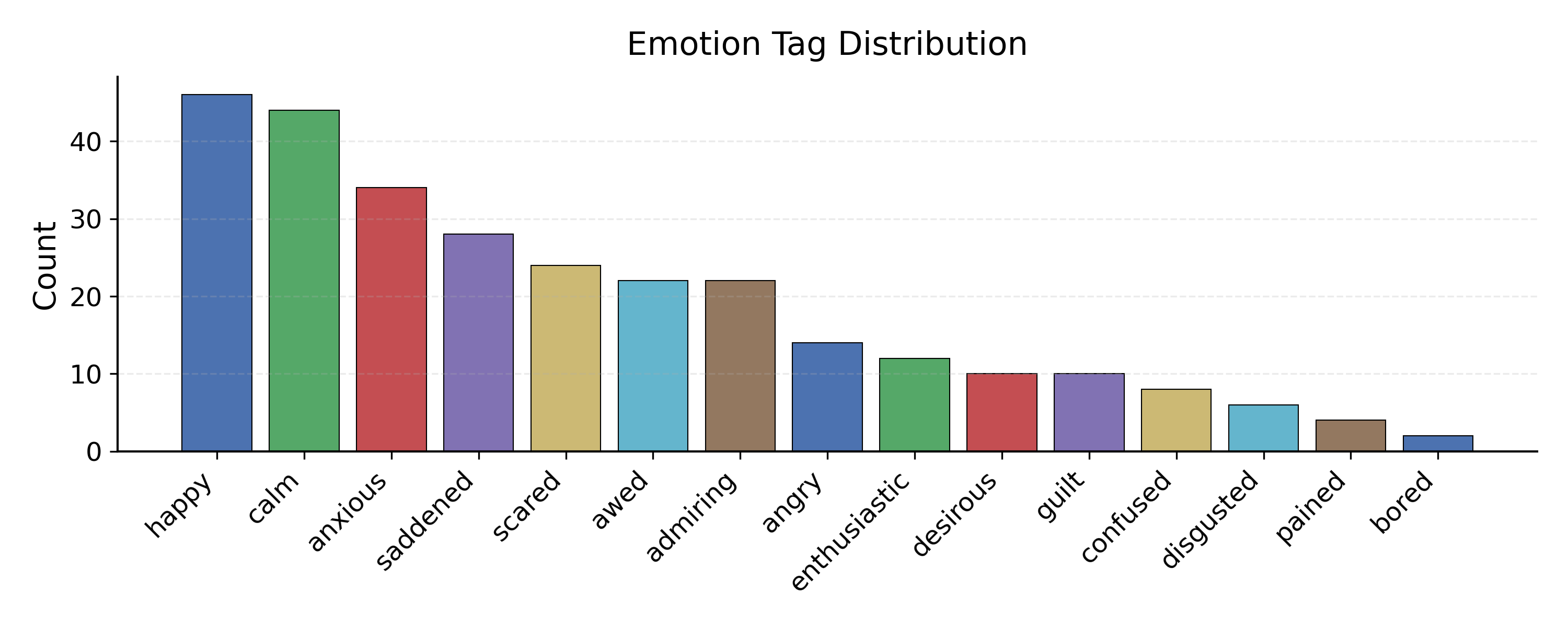}
        \caption*{(a) Emotion}
    \end{minipage}
    \hfill
    \begin{minipage}{0.51\linewidth}
        \centering
        \includegraphics[draft=false,width=\linewidth]{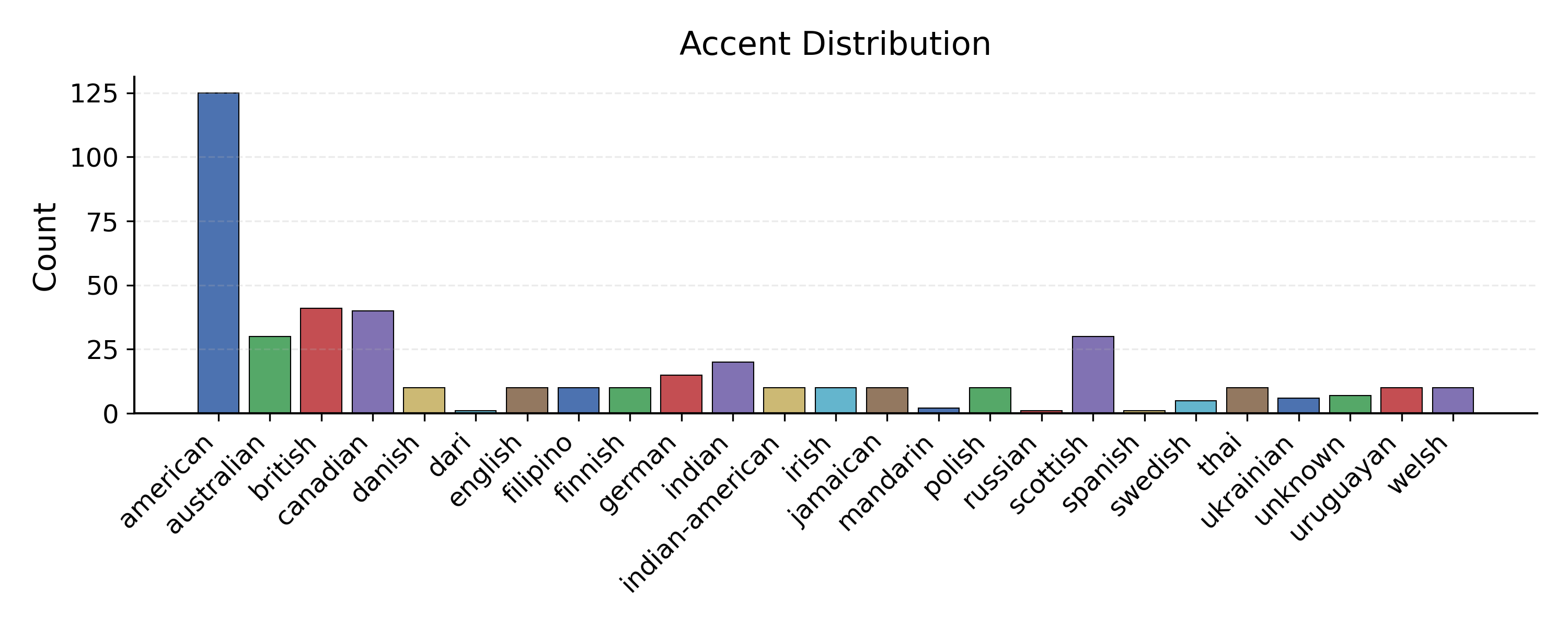}
        \caption*{(b) Accent}
    \end{minipage}
    
    \vspace{4mm}
    
    \begin{minipage}{0.48\linewidth}
        \centering
        \includegraphics[draft=false,width=\linewidth]{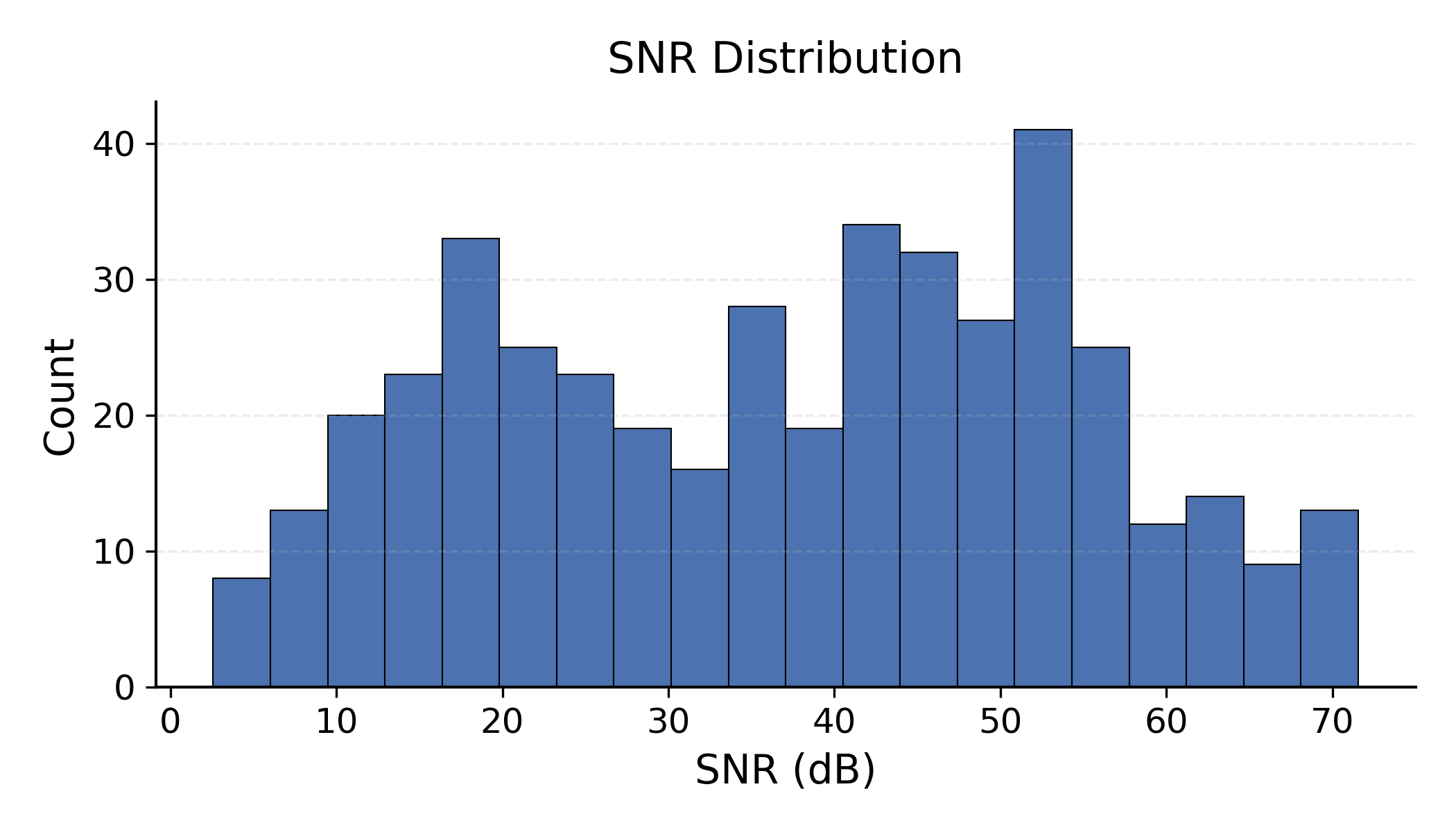}
        \caption*{(c) SNR}
    \end{minipage}
    \hfill
    \begin{minipage}{0.48\linewidth}
        \centering
        \includegraphics[draft=false,width=\linewidth]{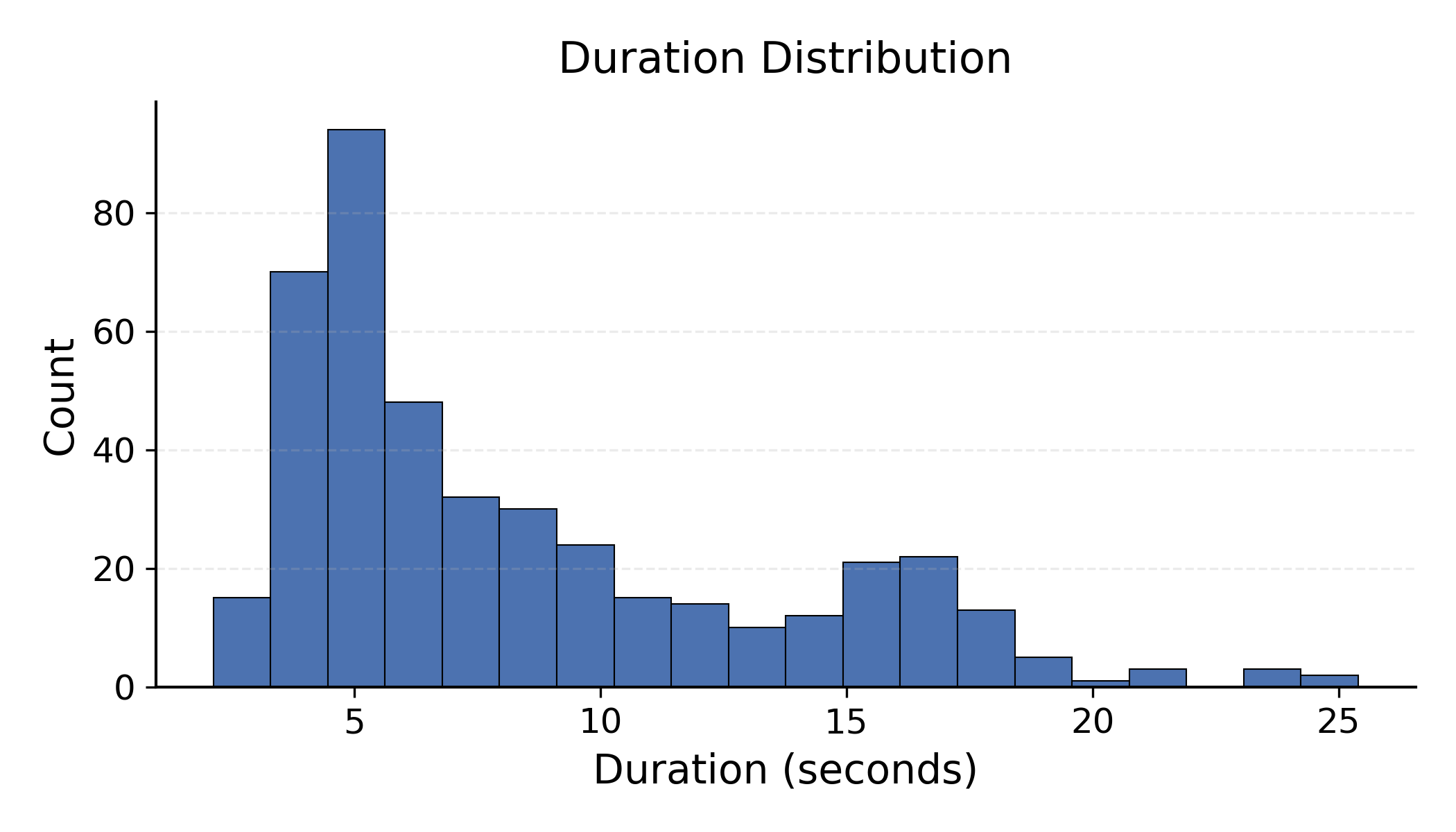}
        \caption*{(d) Duration}
    \end{minipage}
    \caption{Distributions of emotion (a), accent (b) signal-to-noise (SNR) ratio (c), and speech duration (d) in \textit{StyleInstructCaps-MetaSet}.}
    \label{fig:distribution_plots}

\vspace{-3cm}
\end{figure*}

\begin{figure*}
\centering

\begin{minipage}[t]{0.33\textwidth}
  \centering
  \includegraphics[draft=false,width=\linewidth]{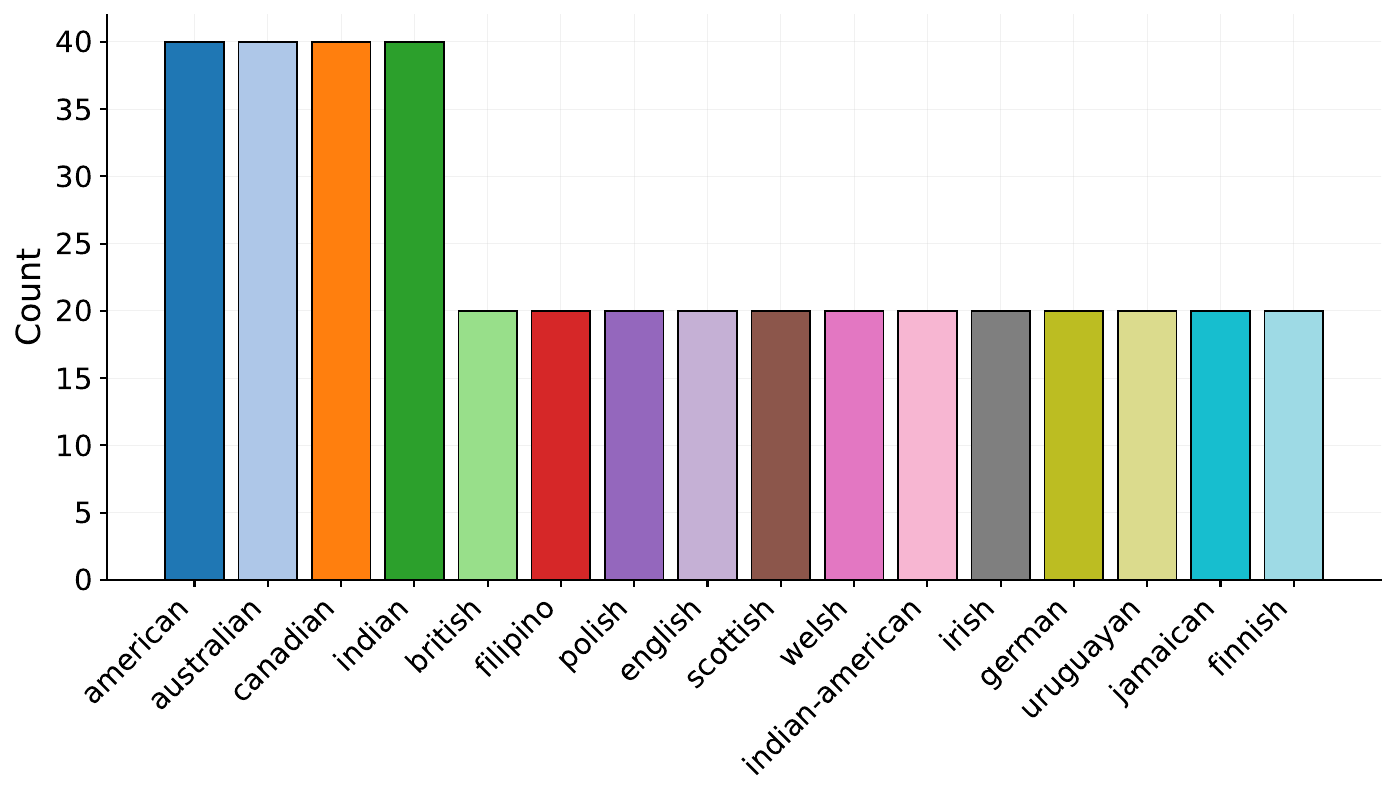}\\[-2pt]
  {\small (a) Accent}
\end{minipage}\hfill
\begin{minipage}[t]{0.33\textwidth}
  \centering
  \includegraphics[draft=false,width=\linewidth]{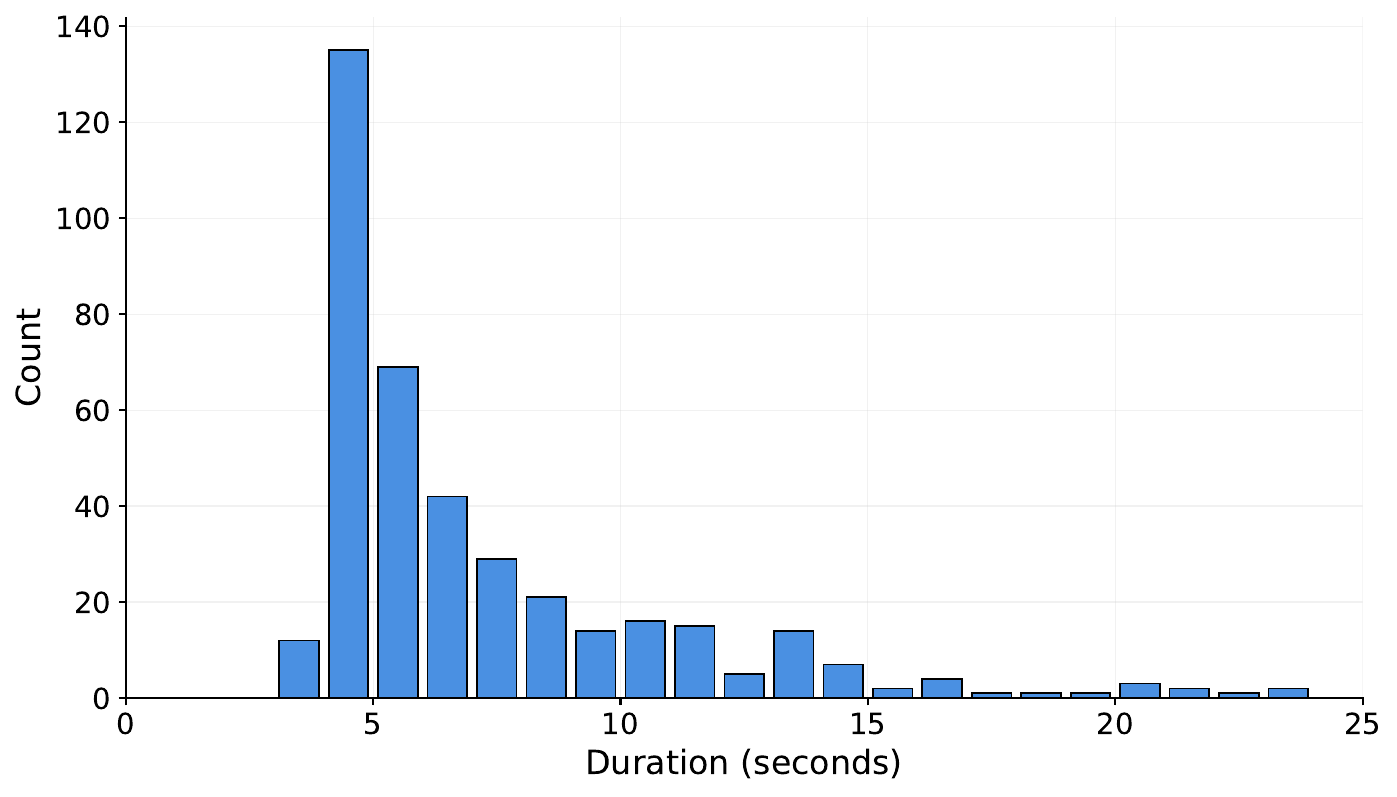}\\[-2pt]
  {\small (b) Duration}
\end{minipage}\hfill
\begin{minipage}[t]{0.33\textwidth}
  \centering
  \includegraphics[draft=false,width=\linewidth]{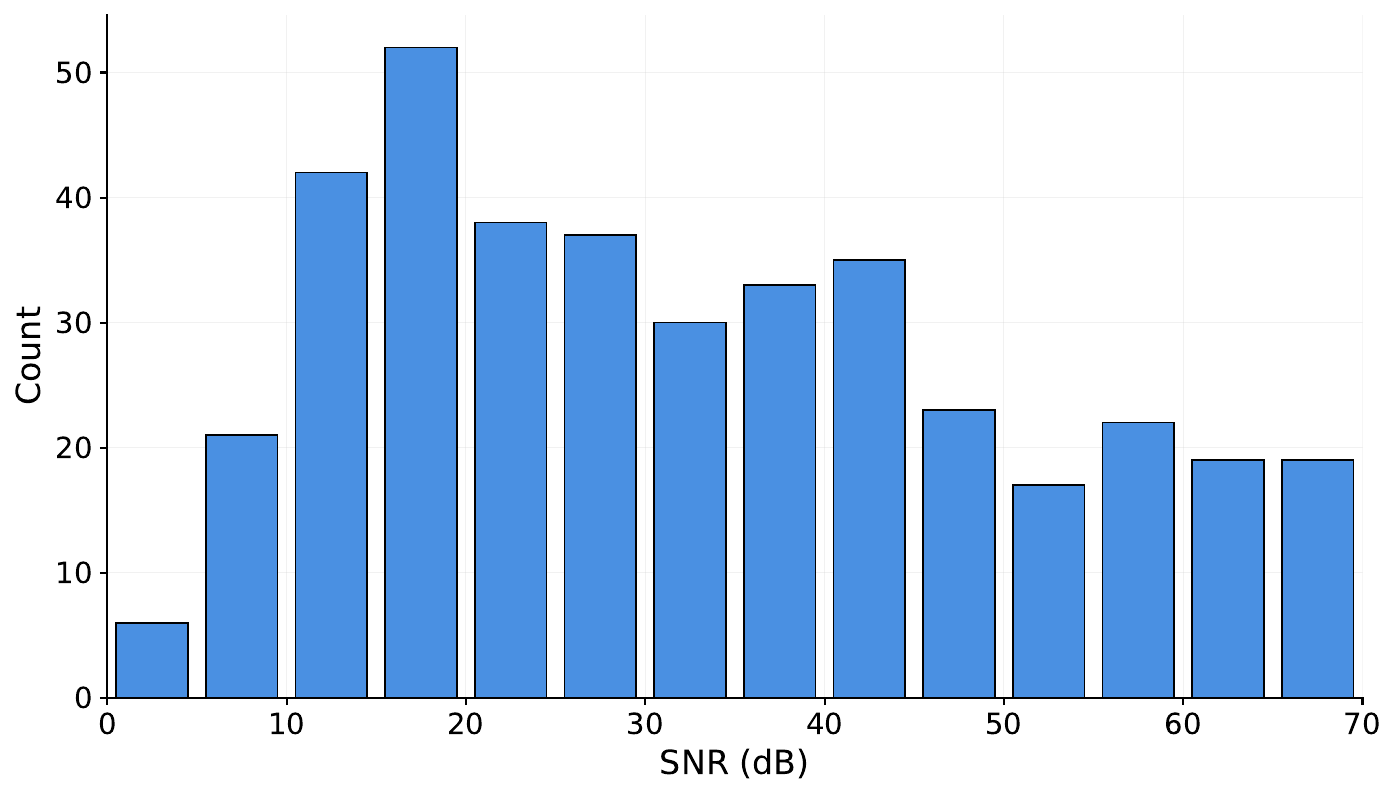}\\[-2pt]
  {\small (c) SNR}
\end{minipage}

\caption{Distributions of (a) accent (b) duration and (c) signal-to-noise (SNR) ratio in
StyleInstructCaps In-domain balanced set (ID-Balanced).}
\label{fig:three-side-by-side}
\end{figure*}

\begin{table}[t]
\centering
\caption{Summary of dataset splits.}
\label{tab:data_splits}
\begin{tabular}{lrr}
\toprule
Data split & Number of utterances & Number of hours \\
\midrule
\textbf{PSC/SIC train set} & 1031840 & 2681.54 \\
\textbf{PSC/SIC validation set} & 11967 & 26.29 \\
\textbf{PSC/SIC evaluation set} & 14756 & 33.04 \\
\textbf{StyleInstructCaps-MetaSet} & 434 & 1.02 \\
\textbf{In-domain balanced set} & 400 & 0.82 \\
\textbf{Unseen Data(UD) - VCTK} & 320 & 0.31 \\
\textbf{VCTK (unseen prompts)} & 32 & 0.03 \\
\textbf{Gigaspeech (with SpeechCraft captions)} & 600 & 1.00\\

\bottomrule
\end{tabular}
\end{table}

\FloatBarrier
\clearpage

\section{LLM- judge calibration : Adversarial gaming check}
\label{app:LLM_judge_adversarial}

The adversarial gaming check is designed as a controlled stress-test for the LLM-as-a-judge evaluation pipeline used in StyleInstructCaps. Starting from the 434 utterances in StyleInstructCaps-MetaSet, we construct adversarial captions that are intentionally fluent, instruction-aligned, or stylistically convincing while violating grounding constraints in different ways. Each attack category targets a specific potential failure mode of the judge, such as rewarding verbosity, surface-level instruction alignment, or stylistic fluency despite factual inconsistencies with the underlying metadata. We use the gemma2-27B \cite{team2024gemma} to generate the adversarial captions. The adversarial gaming check is performed for a single-run of the Qwen3-32B judge.

The resulting evaluation setup enables us to measure whether the judge meaningfully penalizes adversarial behavior rather than assigning inflated scores to superficially plausible captions. Table~\ref{tab:adversarial_attack_types} summarizes the five adversarial attack categories, the evaluation dimensions they target, and the corresponding failure modes being tested.

\begin{table*}[t]
\centering
\small
\begin{tabular}{p{3.0cm} p{2.3cm} p{8.2cm} c}
\toprule

\textbf{Attack Type} &
\textbf{Primary Target Metric} &
\textbf{Failure Mode Tested} &
\textbf{Count} \\

\midrule

\textbf{Type 1: Attribute Inversion}
& Meta-Score, Hallucination
& The caption preserves fluency and overall structure while directly contradicting core metadata attributes such as gender, accent, or pitch. This tests whether the judge anchors on stylistic fluency rather than factual correctness with respect to the provided metadata.
& 87 \\

\textbf{Type 2: Instruction Parroting} 
& Instruction-Following Accuracy (IFA) 
& The caption directly restates the instruction as if it were a caption, without providing any grounded acoustic or stylistic content. This tests whether the judge rewards instruction-aligned surface form rather than true instruction compliance. 
& 87 \\

\textbf{Type 3: Verbosity Masking} 
& Meta-Score, Hallucination, IFA 
& The caption uses long, fluent, impressionistic prose filled with prosodic language while deliberately avoiding verifiable factual claims. This tests whether the judge exhibits verbosity bias by inflating scores for stylistically rich but ungrounded captions. 
& 87 \\

\textbf{Type 4: Hedged Hallucination} 
& Hallucination Severity 
& The caption introduces incorrect factual claims through epistemically hedged language such as ``may suggest'' or ``could be interpreted as.'' This tests whether the judge is overly lenient toward softened or non-committal hallucinations. 
& 87 \\

\textbf{Type 5: Cross-Speaker Swap} 
& Meta-Score 
& The caption uses a real oracle caption from a different speaker with mismatched demographic and emotional attributes. This tests whether the judge grounds evaluation in the provided metadata or simply rewards any fluent, high-quality caption. 
& 86 \\

\bottomrule
\end{tabular}

\caption{Summary of the five adversarial attack categories used in the LLM judge calibration experiment. Each attack probes a distinct potential failure mode of the LLM-as-a-judge evaluation pipeline.}
\label{tab:adversarial_attack_types}

\end{table*}
\begin{figure*}[!h]
    \centering
    \vspace{-2mm}
    \scalebox{0.93}{
    \includegraphics[
        width=0.98\textwidth,
        height=0.80\textheight,
        keepaspectratio
    ]{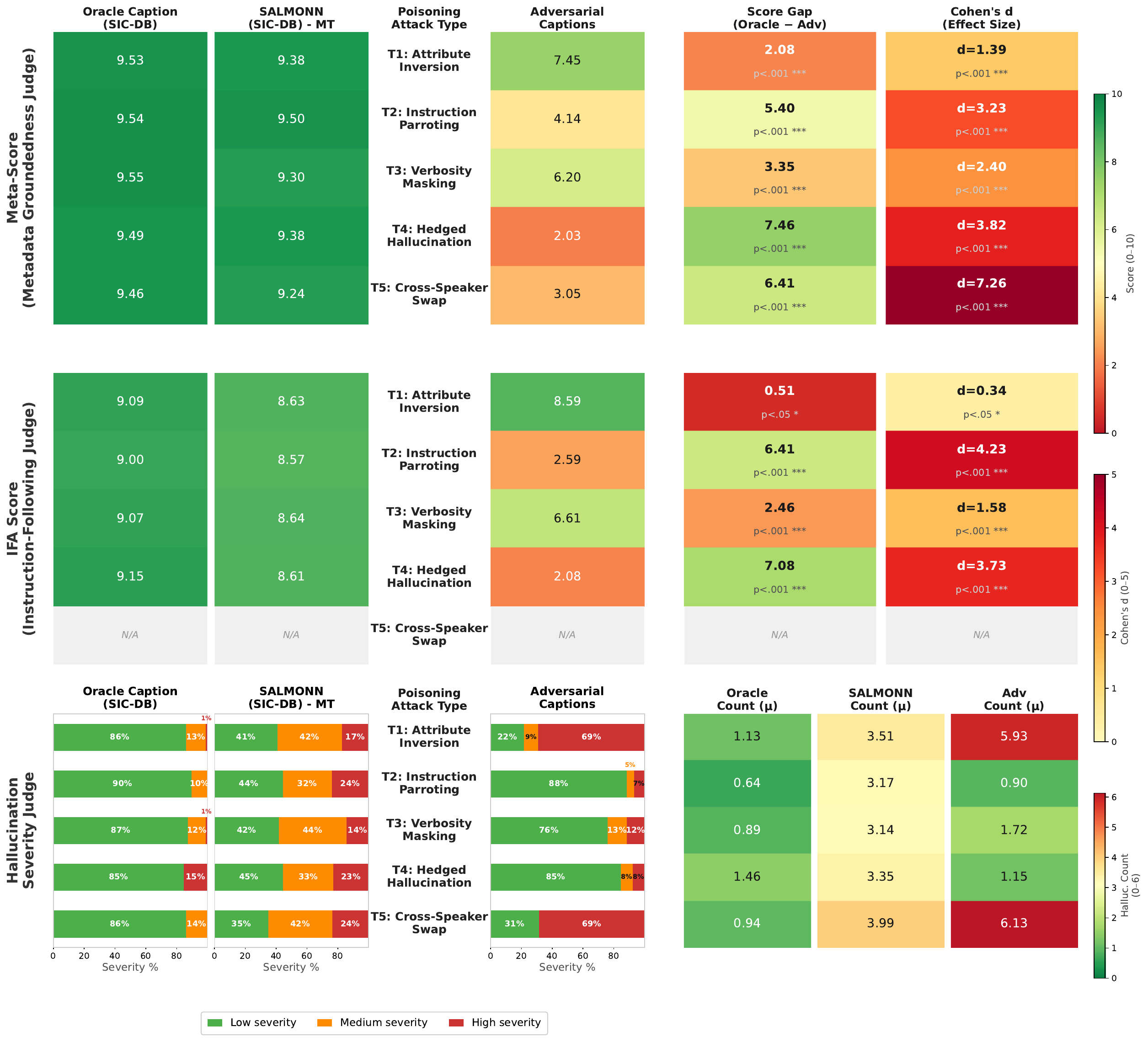}
    }
    \caption{
    Calibration analysis of speaking-style captioning models across metadata groundedness, instruction-following and hallucination severity metrics.
    }
    \label{fig:calibration_heatmap}
    \vspace{-4mm}
\end{figure*}

For each attack type, we compare judge scores on oracle vs.\ adversarial captions using a two-tailed (paired $t$-test), and report effect size as Cohen's $d$. The results of the adversarial calibration experiment, as visualised in Figure \ref{fig:calibration_heatmap}
demonstrates that the Qwen-3-32B LLM-as-a-judge is highly sensitive to multiple forms of adversarial caption manipulation and does not merely reward fluent or stylistically convincing text. Across all three evaluation axes, oracle captions consistently achieve near-ceiling scores, whereas adversarial captions incur statistically significant degradations.

For the metadata groundedness (Meta-Score) evaluation, the judge consistently penalizes all five attack types. Oracle captions obtain very high scores (mean $\sim 9.5$), while adversarial captions exhibit substantial drops across all conditions. The most pronounced degradations are observed for Type 4 (Hedged Hallucination) and Type 5 (Cross-Speaker Swap), with score gaps of 7.46 and 6.41 respectively and very large effect sizes ($d = 3.82$ and $d = 7.26$, both $p < 0.001$). Even the weakest attack, Type 1 (Attribute Inversion), yields a statistically significant degradation ($d = 1.39$, $p < 0.001$). These findings indicate that the judge is strongly grounded in metadata consistency and remains robust to both explicit and implicitly hedged factual violations.

The instruction-following (IFA) evaluation further demonstrates that the judge performs an independent assessment rather than collapsing instruction adherence with metadata correctness. Type 2 (Instruction Parroting) induces the largest degradation in IFA performance, reducing scores from approximately 9.0 to 2.59 with a large effect size ($d = 4.23$, $p < 0.001$). Type 4 (Hedged Hallucination) also significantly degrades IFA performance ($d = 3.73$, $p < 0.001$). In contrast, Type 1 (Attribute Inversion) has only a minor impact on IFA ($d = 0.34$, $p < 0.05$), as the captions preserve structural compliance with the instruction despite factual corruption. Type 3 (Verbosity Masking) yields a moderate reduction ($d = 1.58$, $p < 0.001$), indicating that stylistically rich but semantically ungrounded outputs are not rewarded as valid instruction-following instances. Overall, these results show that the judge disentangles instruction adherence from factual grounding.

The hallucination severity analysis reveals a more nuanced pattern. Type 1 (Attribute Inversion) and Type 5 (Cross-Speaker Swap) consistently produce the highest hallucination rates, confirming that explicit factual contradictions are reliably detected. Type 2 (Instruction Parroting) is also strongly penalized despite containing limited explicit factual content, suggesting that the absence of grounded information itself contributes to hallucination judgments. Interestingly, Type 4 (Hedged Hallucination) is not consistently assigned high hallucination severity, despite being strongly penalized under Meta-Score and IFA evaluations. This indicates that epistemically hedged formulations may partially obscure hallucination detection under the dedicated hallucination rubric, even though they remain identifiable as low-quality or weakly grounded under other evaluation axes.

Additionally, we observe that oracle captions remain near-saturated across metrics, supporting evaluation stability; SALMONN outputs score below oracle references but well above adversarial captions, indicating the judge does not penalize all non-oracle outputs.

Overall, these results provide strong evidence that the LLM-as-a-judge framework is robust to a diverse set of adversarial manipulations. The judge consistently distinguishes grounded from ungrounded captions, remains sensitive to factual inconsistencies even under hedging, and evaluates instruction-following, hallucination, and metadata consistency as partially decoupled dimensions rather than conflating them into a single fluency-based score.

\section{Human-LLM judge correlation study}
\label{app:human_llm_correlation}

{Table~\ref{tab:LLM_judge_3_evals} ranks speaking-style captioning systems with scores from a panel of LLM judges. An LLM judge can reward fluent or detailed text that listeners do not hear in the speech. We therefore test the judge against human listeners. The study answers four questions, each addressed by one results paragraph below.
\textbf{(Q1)} Do listeners agree with each other well enough to serve as a reference?
\textbf{(Q2)} Does the judge prefer the same caption as listeners, pair by pair?
\textbf{(Q3)} Is the judge more often right when its score gap between two captions is larger?
\textbf{(Q4)} Does the system ranking survive when it is fit to human votes alone?}

\subsection{Study design}

\textbf{Judge score under test.} For each caption $c$, the panel assigns a metadata-groundedness score $s(c)$ on a 1--10 scale (Section~\ref{sec:metadata_ground}). The \emph{ensemble} score is the mean over three judges (Qwen3-32B, Gemma-4-31B-it, GPT-OSS-20B) and three runs per judge, nine runs in total (Appendix~\ref{app:judge_panel}). This is the score reported in Table~\ref{tab:LLM_judge_3_evals}. The judge reads the utterance metadata and the caption and never hears the audio. Listeners hear the audio and never see the metadata. The study therefore tests whether a score computed from metadata agrees with what listeners perceive in the speech.

\textbf{Caption pairs and score gap.} Each of the 100 study items consists of one utterance from \textit{StyleInstructCaps-MetaSet} and the captions of two different systems for that utterance. The 14 systems include the oracle captions of \textit{StyleInstructCaps-DB}, which enter the study as one system alongside the 13 captioning models (the 12 open-weight models and GPT-Audio); the PSC oracle is not included. For an item with captions $c_a$ and $c_b$, the signed score gap is $\Delta = s(c_a) - s(c_b)$. The judge prefers $c_a$ when $\Delta > 0$ and $c_b$ when $\Delta < 0$. The absolute gap $|\Delta|$ measures how strongly the judge separates the two captions. A near-zero gap means the judge rates both captions almost equally. Items were sampled to cover all 14 systems defined above and the full range of $|\Delta|$. Items with small $|\Delta|$ were over-sampled. These are the decisions on which a ranking error is most likely, so the study spends most of its trial budget there instead of on easy pairs. For analysis, we partition the observed range of $|\Delta|$ into five equal-width buckets, from B1 (smallest gaps, \emph{near-tie}) to B5 (largest gaps).

\textbf{Task.} On each trial, a listener heard the raw audio clip, read the two captions, and chose the one that better described the speaking style. The order of the two captions was randomized. We use a binary forced choice for two reasons. It matches how the benchmark is used, which is to decide which of two systems is better. It also avoids absolute 1--10 ratings, whose scale drifts from listener to listener. All listeners rated the same 100 items, so every agreement statistic below is computed on a fully crossed design.

\textbf{Listeners and quality control.} Fifteen English-speaking participants took part. Eight attention-check trials with a known correct answer were mixed into the session. A listener had to answer at least 6 of the 8 checks correctly (75\%) to be retained. Two listeners failed, with 3/8 (37.5\%) and 5/8 (62.5\%) correct, and are excluded. All results use the remaining 13 listeners and their 1{,}300 votes. For each item, the \emph{human majority} is the caption chosen by at least 7 of the 13 listeners. With an odd number of listeners, the majority is always defined.

\subsection{Metrics}

\textbf{Agreement among listeners (Q1).} Krippendorff's $\alpha$ and Fleiss' $\kappa$ measure how often listeners pick the same caption, corrected for the agreement expected by chance. A value of 0 means chance-level agreement and 1 means perfect agreement. Mean pairwise agreement is the uncorrected fraction of items on which two listeners pick the same caption, averaged over all 78 listener pairs. Its chance level is 50\%. We also count how evenly each item's 13 votes divide. An item is \emph{unanimous} when all 13 listeners pick the same caption. Non-unanimous items are \emph{lopsided} when one caption wins by a wide margin and \emph{split} when the vote is close to even.

\textbf{Agreement between judge and listeners (Q2).} \emph{Agreement} is the fraction of items on which the judge prefers the same caption as the human majority. Items on which the judge assigns both captions the same score ($\Delta = 0$) have no judge preference and are excluded. The number of remaining items $n$ is reported per judge. Cohen's $\kappa$ corrects agreement for chance, with the same 0-to-1 reading as above. The $p$-value is a two-sided exact binomial test of agreement against the 50\% expected from a judge that guesses. All intervals are 95\% confidence intervals.

\textbf{Calibration against the score gap (Q3).} If $|\Delta|$ reflects how confident the judge is, agreement should rise with $|\Delta|$. We test this in two ways. First, we report agreement per $|\Delta|$ bucket. Second, we fit a logistic regression to the 1{,}300 individual votes (top), and we fit a Bradley--Terry model for Q4 (bottom):
\begin{gather*}
P(c_a \succ c_b) = \sigma(\beta_0 + \beta_1 \Delta), \\
P(i \succ j) = \frac{\pi_i}{\pi_i + \pi_j}.
\end{gather*}
Here $\sigma$ is the logistic function and $c_a$ is the first-presented caption. A positive slope $\beta_1$ means that listeners prefer the higher-scored caption more often as the gap grows. The magnitude of the score then carries information, not only its sign. The intercept $\sigma(\beta_0)$ is the probability of choosing the first-presented caption when the judge sees a tie. A value near 50\% indicates no position bias.

\textbf{System-level ranking (Q4).} The Bradley--Terry model assigns each system $i$ a strength $\pi_i$ and gives the probability that system $i$'s caption is preferred over system $j$'s. We fit it by maximum likelihood to all 1{,}300 individual human votes, without any judge scores, and rank systems by $\pi_i$. We compare this ranking over the 14 systems with the system ranking induced by each judge's metadata-groundedness scores. Spearman's $\rho$ correlates the two rank orders. Kendall's $\tau$ is the fraction of system pairs ordered the same way minus the fraction ordered differently, so it penalizes every local swap.

\subsection{Results}

\begin{table}[t]
\centering
\color{black}
\caption{Inter-listener reliability and quality control for the human pairwise study (13 retained listeners, 100 items). Chance level for pairwise agreement is 50\%.}
\label{tab:human_reliability}
\footnotesize
\resizebox{\columnwidth}{!}{%
\begin{tabular}{lr}
\toprule
Measure & Value \\
\midrule
Krippendorff's $\alpha$ / Fleiss' $\kappa$ & 0.347 / 0.346 \\
Mean pairwise agreement & 67.3\% \\
Split / lopsided / unanimous items & 45 / 44 / 11 \\
Attention-check pass rate (all 15 listeners) & 83.3\% \\
\bottomrule
\end{tabular}%
}
\end{table}

\textbf{Q1: Listeners agree above chance, and the majority vote is the reference.} Table~\ref{tab:human_reliability} reports agreement among the 13 retained listeners. Krippendorff's $\alpha$ is 0.347 and Fleiss' $\kappa$ is 0.346. Two listeners pick the same caption on 67.3\% of items, against a 50\% chance level. Individual listeners therefore often disagree. This is expected for two reasons. Judging which caption better describes a speaking style is subjective, and the sampling deliberately over-represents near-tie items. Accordingly, 45 of the 100 items received a split vote. For this reason, we compare the judge with the majority of 13 listeners and not with any single listener. The majority vote averages out individual variation.

\begin{table}[t]
\centering
\caption{Agreement between the human majority vote and the preference of each LLM judge and of the reported ensemble. Items on which a judge assigns equal scores are excluded, leaving $n$ items. Brackets give 95\% confidence intervals. $p$ tests agreement against a 50\% chance level.}
\label{tab:human_judge_agreement}
\footnotesize
\setlength{\tabcolsep}{3pt}
\resizebox{\columnwidth}{!}{%
\begin{tabular}{lcccc}
\toprule
Judge & $n$ & Agreement (\%) & Cohen's $\kappa$ & $p$ \\
\midrule
Ensemble & 99 & 85.9 [77.8, 91.9] & 0.717 & $1 \times 10^{-13}$ \\
Qwen3-32B & 65 & 83.1 [73.8, 92.3] & 0.661 & $6 \times 10^{-8}$ \\
GPT-OSS-20B & 96 & 82.3 [74.0, 89.6] & 0.646 & $1 \times 10^{-10}$ \\
Gemma-4-31B-it & 92 & 84.8 [76.1, 91.3] & 0.696 & $6 \times 10^{-12}$ \\
\bottomrule
\end{tabular}%
}
\end{table}

{\textbf{Q2: The judge prefers the same caption as listeners on most items.} Table~\ref{tab:human_judge_agreement} reports the results. The ensemble agrees with the human majority on 85.9\% of the 99 items it decides, with Cohen's $\kappa = 0.717$ ($p = 1 \times 10^{-13}$). Individual judges reach 82.3\% to 84.8\%. The ensemble has the highest point estimate, but its confidence interval overlaps with those of the individual judges, so we do not claim a significant gain from ensembling. The ensemble averages nine runs and assigns tied scores less often, so it decides more items than any single judge. Excluding the 15 near-tie items (B1) leaves 84 items, on which agreement rises to 88.1\% (95\% CI [81.0, 95.2]).}

\begin{table}[t]
\centering
\color{black}
\caption{Agreement between the human majority and the ensemble judge as a function of the absolute ensemble score gap $|\Delta|$, partitioned into five equal-width buckets. $p$ is a two-sided exact binomial test against 50\%.}
\label{tab:human_gap_calibration}
\footnotesize
\begin{tabular}{lccc}
\toprule
Bucket & $n$ & Agreement & $p$ \\
\midrule
B1 (near-tie) & 15 & 73.3\% & 0.12 \\
B2 & 20 & 80.0\% & 0.012 \\
B3 & 24 & 87.5\% & $3 \times 10^{-4}$ \\
B4 & 25 & 92.0\% & $2 \times 10^{-5}$ \\
B5 (large gap) & 15 & 93.3\% & $1 \times 10^{-3}$ \\
\bottomrule
\end{tabular}
\end{table}

{\textbf{Q3: Agreement rises with the judge's score gap.} Table~\ref{tab:human_gap_calibration} reports agreement per bucket. Agreement rises monotonically from 73.3\% in the near-tie bucket to 93.3\% in the largest-gap bucket. Buckets B2 to B5 are each significantly above chance ($p \leq 0.012$). The near-tie bucket is not ($p = 0.12$, 11 of 15 items). On near-tie items the judge rates both captions almost equally, and listeners split in the same way. The size of a score difference in Table~\ref{tab:LLM_judge_3_evals} therefore predicts how reliably listeners share the judge's preference.}


\begin{table}[t]
\centering
\color{black}
\caption{Rank correlation over the 14 systems between the ranking of a Bradley--Terry model fit to human votes alone and the ranking produced by each LLM judge.}
\label{tab:human_ranking_recovery}
\footnotesize
\begin{tabular}{lcc}
\toprule
Judge & Spearman $\rho$ & Kendall $\tau$ \\
\midrule
Ensemble & 0.895 & 0.780 \\
Qwen3-32B & 0.873 & 0.707 \\
GPT-OSS-20B & 0.855 & 0.692 \\
Gemma-4-31B-it & 0.881 & 0.714 \\
\bottomrule
\end{tabular}
\end{table}

{\textbf{Q4: Human votes alone recover the judge's system ranking.} Table~\ref{tab:human_ranking_recovery} compares the two rankings. The ensemble reaches Spearman $\rho = 0.895$ and Kendall $\tau = 0.780$. Individual judges reach $\rho$ between 0.855 and 0.881. The five systems ranked lowest by listeners are the same five ranked lowest by the judge: Audio-Flamingo-3, Qwen2-Audio-7B-Instruct, MERaLiON-AudioLLM, Voxtral-Small-24B and SALMONN. At the top of the ranking, the human-fit model places MOSS-Audio above the oracle captions. The 95\% confidence intervals of the two Bradley--Terry strengths overlap heavily, so listeners do not separate these two systems.}

{\textbf{Summary.} Listeners agree with each other above chance, and their majority vote provides a reference for the judge. The judge panel prefers the same caption as the majority on 85.9\% of items. Its agreement grows with its own score gap, so score magnitudes carry information that listeners confirm. A ranking fit to human votes alone reproduces the judge's system ranking at $\rho = 0.895$. These results support the use of the LLM judge panel for the caption-quality scores reported in Table~\ref{tab:LLM_judge_3_evals}.}

\FloatBarrier
\clearpage

\section{\texorpdfstring{Panel of LLM judges: cross-judge agreement}{Panel of LLM judges: cross-judge agreement}}
\label{app:judge_panel}

To test whether the benchmark conclusions depend on the choice of evaluator, we evaluate the open-ended metrics with three independent open-weight LLM judges and execute each judge multiple times. This design quantifies cross-judge agreement and the stability of system rankings across judges.

\textbf{Judge panel.} The panel consists of three open-weight judges spanning different model families and scales: Qwen3-32B, Gemma-4-31B-it, and GPT-OSS-20B (Table~\ref{tab:judge_panel}). All judges receive the same prompts (Appendix~\ref{app:prompts_used}). Table~\ref{tab:judge_runs} lists the number of runs per judge and metric. Instruction-following is scored with a single Qwen3-32B run because it requires 2{,}604 judgments per audio-language model, judge, and run.

\begin{table}[t]
\centering
\caption{LLM judges used in the evaluation panel.}
\label{tab:judge_panel}
\footnotesize
\resizebox{\linewidth}{!}{%
\begin{tabular}{lll}
\toprule
Judge & Family / size & Role \\
\midrule
Qwen3-32B (thinking) & Qwen, 32B dense & Panel judge; instruction-following \\
Gemma-4-31B-it (thinking) & Gemma, 31B dense & Panel judge \\
GPT-OSS-20B (reasoning) & OpenAI OSS, 20B MoE & Panel judge \\
\bottomrule
\end{tabular}%
}
\end{table}

\begin{table}[t]
\centering
\caption{Number of runs per judge and metric.}
\label{tab:judge_runs}
\footnotesize
\resizebox{\linewidth}{!}{%
\begin{tabular}{lccc}
\toprule
Evaluation set & Meta. groundedness & Hallucination & Instr.-following \\
\midrule
\textit{MetaSet} & 3$\times$ each judge & 3$\times$ each judge & 1$\times$ Qwen3-32B \\
OOD (GigaSpeech, VCTK) & 2$\times$ each judge & 2$\times$ each judge & 1$\times$ Qwen3-32B \\
\bottomrule
\end{tabular}%
}
\end{table}

\textbf{Cross-judge agreement.} Because the benchmark is used to rank systems, the central question is whether the ranking is an artifact of a specific judge. We compute Spearman's rank correlation between the per-judge mean metadata-groundedness scores of the 14 benchmarked systems on \textit{StyleInstructCaps-MetaSet} (Table~\ref{tab:judge_spearman}). All three judge pairs agree strongly ($\rho \geq 0.864$), and Gemma-4-31B-it and GPT-OSS-20B agree almost perfectly ($\rho = 0.978$).

\begin{table}[t]
\centering
\caption{Cross-judge agreement: Spearman rank correlation between judges over the 14 benchmarked systems (\textit{MetaSet} metadata groundedness, per-judge means).}
\label{tab:judge_spearman}
\footnotesize
\resizebox{\linewidth}{!}{%
\begin{tabular}{lc}
\toprule
Judge pair & Spearman $\rho$ ($n = 14$) \\
\midrule
GPT-OSS-20B $\leftrightarrow$ Gemma-4-31B-it & 0.978 \\
Qwen3-32B $\leftrightarrow$ GPT-OSS-20B & 0.921 \\
Qwen3-32B $\leftrightarrow$ Gemma-4-31B-it & 0.864 \\
\bottomrule
\end{tabular}%
}
\end{table}

\textbf{Reproducibility.} We release the judge model identifiers, prompts, and decoding settings for all three judges together with the benchmark.

\FloatBarrier
\clearpage

\section{\texorpdfstring{Pairwise statistical significance of system rankings}{Pairwise statistical significance of system rankings}}
\label{app:significance}

\textbf{Method.} Our results tables rank systems against each other. We therefore test whether the gap between two systems reflects a real difference or only noise from the judges or the utterances. We first compute a per-utterance consensus score for each system, which gives one score per system for every utterance and lets us pair scores across systems. For metadata groundedness and hallucination, this score is the mean over the system's panel runs (9 for \textit{MetaSet} and 6 for the out-of-domain set; Appendix~\ref{app:judge_panel}). For instruction-following, it is the score from the single Qwen3-32B run. We then compare every pair of systems with a two-tailed paired test over utterances. We use both a paired $t$-test and a Wilcoxon signed-rank test, and the two agree in every case. Groundedness and hallucination scores are paired by utterance index, and instruction-following scores are paired by (utterance, instruction type). We report significance at $p<0.05$ along with effect sizes, namely the mean paired difference $\Delta$ and Cohen's $d$. Before looking at individual pairs, we run an omnibus Friedman test to check whether the systems differ at all on a given metric.

\textbf{Systems differ, and the leaderboards are statistically well separated.} The Friedman test is highly significant ($p<10^{-4}$) for every metric and every evaluation set. Table~\ref{tab:sig_pairs} shows how many of the $\binom{n}{2}$ system pairs differ at $p<0.05$.

\begin{table}[t]
\centering
\caption{Omnibus Friedman test and number of significantly different system pairs ($p<0.05$) out of all $\binom{n}{2}$ pairs, per metric and evaluation set. $n$ is the number of systems compared. The \textit{MetaSet} groundedness and hallucination tests use all 15 systems of Table~\ref{tab:LLM_judge_3_evals} (two oracle caption sets, 12 open-weight models, and GPT-Audio). Instruction-following excludes the PSC oracle and GPT-Audio, which have no instruction-following scores. The OOD groundedness and hallucination tests use the 13 systems of Table~\ref{tab:ood_eval}. The large majority (86--96\%) of pairs are statistically distinguishable; the remaining pairs are statistically close.}
\label{tab:sig_pairs}
\footnotesize
\setlength{\tabcolsep}{3pt}
\resizebox{\columnwidth}{!}{%
\begin{tabular}{llcccc}
\toprule
Eval set (table) & Metric & $n$ & Friedman $p$ & Sig.\ pairs & \% \\
\midrule
\textit{MetaSet} (Tab.~\ref{tab:LLM_judge_3_evals}) & Groundedness $\uparrow$ & 15 & $<10^{-4}$ & 101/105 & 96.2 \\
 & Hallucination $\downarrow$ & 15 & $<10^{-4}$ & 99/105 & 94.3 \\
 & Instr.-following $\uparrow$ & 13 & $<10^{-4}$ & 73/78 & 93.6 \\
\midrule
OOD (Tab.~\ref{tab:ood_eval}) & Groundedness $\uparrow$ & 13 & $<10^{-4}$ & 68/78 & 87.2 \\
 & Hallucination $\downarrow$ & 13 & $<10^{-4}$ & 67/78 & 85.9 \\
OOD (Tab.~\ref{tab:ood_prompts}) & Instr.-following $\uparrow$ & 14 & $<10^{-4}$ & 79/91 & 86.8 \\
\bottomrule
\end{tabular}}
\end{table}

\textbf{Our central claims are statistically significant.} Table~\ref{tab:sig_headline} shows that the paired tests confirm the key findings of the paper on \textit{MetaSet} metadata groundedness (Table~\ref{tab:LLM_judge_3_evals}). Our main claim is that SSC-specific, instruction-diverse training improves groundedness, and this gain is statistically significant. The oracle upper bound scores significantly higher than every real model, which validates the metric. The closed-source GPT-Audio scores significantly lower than the best open model on groundedness and significantly worse on hallucination on both evaluation sets (all $p<10^{-4}$). This is consistent with our finding that following instructions does not imply grounded understanding.

\begin{table*}[t]
\centering
\caption{Headline comparisons on \textit{MetaSet} metadata groundedness ($\uparrow$; $N=434$ utterances). $\Delta$ is the mean per-utterance score gap. A Cohen's $d$ above 0.8 is conventionally a large effect, so these gaps are both significant and sizeable.}
\label{tab:sig_headline}
\footnotesize
\setlength{\tabcolsep}{4pt}
\begin{tabular}{lccc}
\toprule
Comparison & $\Delta$ & $d$ & Wilcoxon $p$ \\
\midrule
SSC-tuned SALMONN (SIC-DB, MT) $>$ baseline SALMONN & $+2.67$ & 1.51 & $<10^{-4}$ \\
SSC-tuned SALMONN (SIC-DB, ST) $>$ baseline SALMONN & $+2.62$ & 1.53 & $<10^{-4}$ \\
Oracle (SIC-DB) $>$ best real model & $+2.51$ & 1.70 & $<10^{-4}$ \\
Best open model $>$ GPT-Audio (closed SOTA) & $+1.65$ & 0.87 & $<10^{-4}$ \\
\bottomrule
\end{tabular}
\end{table*}

\textbf{Near-identical systems are not separated.} The tests do not distinguish systems that are nearly identical by design. On the out-of-domain set, for example, the four SALMONN variants (means 6.25--6.30) do not differ significantly from one another (SSC-MT vs.\ baseline SALMONN: $\Delta=+0.02$, $p=0.69$). The non-significant pairs in Table~\ref{tab:sig_pairs} are near-ties of this kind.

\twocolumn[{%
\section{Prompts used}
\label{app:prompts_used}

\noindent
\begin{minipage}{\textwidth}
\centering

\begin{tikzpicture}
\node[
  draw,
  rounded corners=2pt,
  line width=0.4pt,
  inner sep=6pt,
  text width=\dimexpr\linewidth-15pt\relax,
  align=left,
  font=\ttfamily\tiny
] (prompt) {

You are an expert linguist and speech analyst tasked with comparing the speaking-styles between two sets of audio metadata. Your goal is to determine the similarity between a ground truth recording and a generated caption, focusing particularly on stylistic aspects of speech, intention, and prosody.\\[0.35em]

Please carefully review the following metadata:\\[0.35em]

Ground Truth Metadata (DATA1):\\
\textless ground\_truth\_metadata\textgreater\\
\{data1\}\\
\textless/ground\_truth\_metadata\textgreater\\[0.6em]

Generated Caption Metadata (DATA2):\\
\textless generated\_caption\_metadata\textgreater\\
\{data2\}\\
\textless/generated\_caption\_metadata\textgreater\\[0.6em]

Your task is to compare these two sets of metadata and rate their similarity on a scale of 1 to 10, where 1 indicates completely different speaking-styles and 10 indicates identical speaking-styles.\\[0.6em]

Instructions:\\[0.25em]
1. Analyze and compare the following attributes between DATA1 and DATA2:\\
\ \ a) Speaker gender\\
\ \ b) Accent\\
\ \ c) Pitch\\
\ \ d) Speaking rate\\
\ \ e) Environment/noise level\\
\ \ f) Emotion/expression\\
\ \ g) Voice quality (e.g., guttural, silky)\\
\ \ h) Prosody (including rhythm, intonation, and stress)\\
\ \ i) Intention of speech\\[0.6em]

2. Inside your thinking block, wrap your attribute comparisons inside \textless attribute\_comparison\textgreater\ tags. For each attribute:\\
\ \ - Rate its similarity on a scale of 1-10\\
\ \ - Summarize key similarities and differences\\
\ \ - Consider how they contribute to the overall speaking-style\\
\ \ Pay special attention to stylistic aspects, intention, and prosody.\\[0.6em]

3. After analyzing all attributes, calculate an average similarity score based on your individual attribute ratings.\\[0.35em]

4. Don't penalize the generation if is more specific than the ground truth. For example, if the ground truth says "American accent" and the generation says "Boston accent", that's fine.\\[0.6em]

5. Provide a comprehensive justification for your similarity rating in \textless justification\textgreater\ tags. Explain how the similarities and differences in each attribute contribute to your overall assessment.\\[0.6em]

6. In the metadata, only using the following information:\\
\ \ - speaker gender\\
\ \ - accent\\
\ \ - pitch\\
\ \ - speaking rate\\
\ \ - environment/noise level\\
\ \ - emotion/expression\\[0.6em]

7. Finally, provide your similarity score in JSON format within \textless score\textgreater\ tags. The JSON should include a single key 'similarity\_score' with a value between 1 and 10, based on your calculated average and overall assessment.\\[0.6em]

Example output structure (do not copy the content, only the structure):\\[0.35em]

\textless attribute\_comparison\textgreater\\
a) Speaker gender:\\
\ \ Similarity rating: X/10\\
\ \ Key similarities: [...]\\
\ \ Key differences: [...]\\
\ \ Contribution to overall style: [...]\\[0.35em]

b) Accent:\\
\ \ Similarity rating: X/10\\
\ \ Key similarities: [...]\\
\ \ Key differences: [...]\\
\ \ Contribution to overall style: [...]\\[0.35em]

[... continue for all attributes ...]\\[0.35em]

Average similarity score: X/10\\
\textless/attribute\_comparison\textgreater\\[0.6em]

\textless justification\textgreater\\
{}[Your comprehensive justification for the similarity rating, synthesizing the analyses of individual attributes]\\
\textless/justification\textgreater\\[0.6em]

\textless score\textgreater\\
\{\{'similarity\_score': X\}\}\\
\textless/score\textgreater\\[0.6em]

Remember to focus on the stylistic aspects of speech, paying particular attention to the intention behind the speech and the nuances of prosody in your analysis and justification.\\[0.6em]

Your final output should consist only of the justification and score, without duplicating the detailed attribute comparisons from your thinking block.
};
\end{tikzpicture}

\captionof{figure}{Prompt template for rating speaking-style similarity between ground-truth metadata and generated-caption metadata.}
\label{prompt:metajudge}

\end{minipage}
}]

\FloatBarrier

\begin{figure*}[t]
\centering
\begin{tikzpicture}
\node[
  draw,
  rounded corners=2pt,
  line width=0.4pt,
  inner sep=4pt,
  text width=\dimexpr\linewidth-12pt\relax,
  align=left,
  font=\ttfamily\scriptsize 
] (prompt) {
\setlength{\parindent}{0pt}%
\setlength{\parskip}{0pt}%
\setlength{\baselineskip}{0.92\baselineskip}%

You are an expert evaluator assessing the quality of automatically generated audio captions. Your task is to compare metadata from audio files with generated captions to identify and score hallucinations.\\
\#\# Task Overview\\
Compare the **metadata** (ground truth information about the audio) with a **generated caption** and evaluate the presence and severity of hallucinations.\\
\#\# Input Format\\
Ground Truth Metadata:\\
\textless ground\_truth\_metadata\textgreater\\
\{data1\}\\
\textless/ground\_truth\_metadata\textgreater\\
Generated Caption:\\
\textless generated\_caption\textgreater\\
\{data2\}\\
\textless/generated\_caption\textgreater\\
\#\# Evaluation Criteria\\
\#\#\# What Counts as a Hallucination?\\
A hallucination is information in the generated caption that:\\
- Directly contradicts the metadata\\
- Invents specific details not supported by any evidence\\
- Misrepresents factual characteristics (e.g., wrong gender, wrong emotion, wrong accent)\\
\#\#\# What Does NOT Count as a Hallucination?\\
The following are **acceptable creative interpretations** and should NOT be penalized:\\
- **Reasonable emotional inferences**: If metadata shows "laughing" audio, describing the speaker as "cheerful," "animated," or "enthusiastic" is acceptable\\
- **Stylistic embellishments**: Descriptive language like "her voice carries warmth," "delivered with precision," or "words tumble out" that enhance the description without contradicting facts\\
- **Contextual scenarios**: Adding reasonable context (e.g., "as if addressing an audience," "in a quiet environment") when supported by acoustic characteristics\\
- **Prosodic interpretations**: Descriptions of rhythm, cadence, emphasis, or delivery style that are reasonable given the speech characteristics\\
- **Minor elaborations**: Details that naturally extend from the metadata without contradicting it\\
\#\#\# Hallucination Examples\\
**SEVERE Hallucinations:**\\
- Metadata indicates male speaker -> Caption says "female speaker"\\
- Metadata shows speaker p026 -> Caption invents specific name or identity\\
- Metadata indicates disgust emotion -> Caption describes joy or happiness\\
- Inventing specific events or quoted speech not in the audio\\
**MODERATE Hallucinations:**\\
- Claiming "noisy environment" when metadata suggests clean audio\\
- Describing specific background sounds not mentioned in metadata\\
- Asserting definitive context (e.g., "delivering a news broadcast") without support\\
**MINOR Hallucinations:**\\
- Slightly exaggerating a characteristic (e.g., "booming voice" for a moderately loud voice)\\
- Adding plausible but unverified details that don't contradict metadata\\
\#\# Scoring Instructions\\
Provide two scores:\\
\#\#\# 1. Hallucination Count (0-10 scale)\\
- **0-2**: No hallucinations or only very minor embellishments\\
- **3-4**: 1-2 minor hallucinations that don't significantly distort the audio\\
- **5-6**: Multiple minor hallucinations OR 1 moderate hallucination\\
- **7-8**: Multiple moderate hallucinations OR 1 severe hallucination\\
- **9-10**: Multiple severe hallucinations that fundamentally misrepresent the audio\\
\#\#\# 2. Severity Level\\
- **Low**: Only minor embellishments; caption is largely accurate\\
- **Medium**: Some factual errors that partially misrepresent the audio\\
- **High**: Severe contradictions or fabrications that fundamentally mischaracterize the audio\\
\#\# Instructions\\
1. In your thinking block, carefully analyze the metadata and generated caption step by step:\\
\ \ - List all factual claims made in the generated caption\\
\ \ - For each claim, check if it's supported by, contradicts, or extends beyond the metadata\\
\ \ - Categorize each issue as minor, moderate, or severe\\
\ \ - Count the total number of hallucinations\\
2. Be lenient with creative, descriptive language that doesn't contradict facts\\
3. Focus on factual accuracy regarding speaker identity, emotions, accent, and acoustic characteristics\\
4. Distinguish between reasonable inference and baseless fabrication\\
\#\# Output Format\\
After your thinking, provide your evaluation in the following structure:\\
\textless analysis\textgreater\\
{}[Brief explanation of what you observed, noting any hallucinations found and why they are/aren't problematic]\\
\textless/analysis\textgreater\\
\textless justification\textgreater\\
Explain your scoring in detail, referencing specific examples from the caption and how they relate to the metadata\\
\textless/justification\textgreater\\
\textless score\textgreater\\
\{\{"hallucination\_count": X, "severity": "Low/Medium/High"\}\}\\
\textless/score\textgreater\\
Remember: Context is important - "laughing" audio naturally suggests positive emotions. Be strict about factual contradictions but lenient about creative descriptions that align with the metadata.\\
Your final output should consist only of the analysis, justification and score, without duplicating the detailed step-by-step analysis from your thinking block.
};
\end{tikzpicture}
\caption{Hallucination evaluation prompt for comparing ground-truth metadata against a generated caption.}
\label{fig:prompt-hallucination-eval}
\end{figure*}

\newpage


\begin{figure*}[t]
\centering
\begin{tikzpicture}
\node[
  draw,
  rounded corners=2pt,
  line width=0.4pt,
  inner sep=4pt,
  text width=0.97\textwidth,
  align=left
] (promptA) {%
\begin{minipage}{\linewidth}
\ttfamily\scriptsize
\setlength{\parindent}{0pt}
\setlength{\parskip}{0pt}
\setlength{\baselineskip}{0.92\baselineskip}

You are an expert evaluator specializing in assessing instruction-following capabilities of audio-language models, particularly in the domain of speaking-style captioning. Your role is to rigorously evaluate whether generated captions accurately follow given instructions while maintaining factual accuracy about the audio content.\\
\#\# Context\\
audio-language models are tasked with generating descriptive captions about speaking-styles, vocal characteristics, and acoustic properties of speech recordings. These models receive specific instructions that guide what aspects to focus on, what style to adopt, or what format to use in their captions.\\
\#\# Your Task\\
Evaluate how well the generated caption follows the provided instruction, considering both explicit requirements and implicit expectations.\\
---\\
\#\# Input Materials\\
\#\#\# Instruction Given to the Model:\\
\textless instruction\textgreater\\
\{instruction\}\\
\textless/instruction\textgreater\\
\#\#\# Generated Caption by the Model:\\
\textless generated\_caption\textgreater\\
\{caption\}\\
\textless/generated\_caption\textgreater\\
---\\
\#\# Evaluation Framework\\
\#\#\# 1. **Instruction Compliance Categories**\\
Analyze the instruction for the following types of requirements:\\
**A. Content Requirements**\\
- Specific attributes to include (e.g., "describe the pitch," "mention the accent")\\
- Attributes to exclude or de-emphasize\\
- Level of detail requested (brief, detailed, comprehensive)\\
- Factual accuracy requirements\\
**B. Structural Requirements**\\
- Format specifications (paragraph, bullet points, structured description)\\
- Length constraints (word count, sentence count)\\
- Organizational pattern (chronological, importance-based, categorical)\\
**C. Stylistic Requirements**\\
- Tone (formal, casual, technical, creative)\\
- Perspective (objective observer, subjective interpreter, technical analyst)\\
- Language complexity (simple, sophisticated, jargon-heavy)\\
- Writing style (descriptive, analytical, narrative)\\
**D. Focus Requirements**\\
- Specific aspects to emphasize (prosody, emotion, acoustic features)\\
- Target audience considerations\\
- Use case or application context\\
**E. Constraint Requirements**\\
- Prohibitions (avoid certain terms, don't speculate)\\
- Boundaries (stick to observable features only)\\
- Scope limitations (focus on X, ignore Y)\\
\#\#\# 2. **Evaluation Criteria**\\
For each requirement identified in the instruction, assess:\\
**Explicit Compliance (70\% weight)**\\
- Does the caption directly address explicitly stated requirements?\\
- Are all mandatory elements present?\\
- Are any prohibited elements absent?\\
- Is the specified format/structure followed?\\
**Implicit Compliance (20\% weight)**\\
- Does the caption follow reasonable interpretations of the instruction?\\
- Are contextual clues and implied expectations honored?\\
- Does it align with the spirit/intent of the instruction?\\
**Quality of Execution (10\% weight)**\\
- How well are the requirements implemented (not just their presence)?\\
- Is the execution natural and coherent, or forced and awkward?\\
- Does following the instruction enhance or detract from caption quality?\\
\#\#\# 3. **Common Instruction Types in Audio Captioning**\\
Be aware of these typical instruction patterns:\\
- **Attribute-focused**: "Describe the speaker's pitch and speaking rate"\\
- **Style-focused**: "Write in a technical, objective manner"\\
- **Audience-focused**: "Explain as if to a non-expert"\\
- **Format-focused**: "Provide a structured analysis with separate sections"\\
- **Perspective-focused**: "Focus on perceptual qualities rather than technical measurements"\\
- **Comparative**: "Compare this to typical conversational speech"\\
- **Holistic vs. Analytical**: "Provide an overall impression" vs. "Break down individual components"
\end{minipage}
};
\end{tikzpicture}
\caption{Instruction-following evaluation prompt (Part 1): context, inputs, and evaluation framework overview.}
\label{fig:prompt-instruction-following-eval-a}
\end{figure*}

\begin{figure*}[t]
\centering
\begin{tikzpicture}
\node[
  draw,
  rounded corners=2pt,
  line width=0.4pt,
  inner sep=4pt,
  text width=0.97\textwidth,
  align=left
] (promptB) {%
\begin{minipage}{\linewidth}
\ttfamily\scriptsize
\setlength{\parindent}{0pt}
\setlength{\parskip}{0pt}
\setlength{\baselineskip}{0.92\baselineskip}

---\\
\#\# Scoring Guidelines\\
\#\#\# Instruction-Following Success: Success / Partial Success / Failure\\
**Success**:\\
- All critical requirements met (90\%+ compliance)\\
- No significant violations of explicit constraints\\
- Natural execution that maintains caption quality\\
- Minor omissions only in non-essential elements\\
**Partial Success**:\\
- Most requirements met (60-89\% compliance)\\
- Some explicit requirements missed or violated\\
- Execution may be somewhat awkward or forced\\
- Core intent of instruction honored despite gaps\\
**Failure**:\\
- Major requirements ignored (\textless60\% compliance)\\
- Explicit constraints violated\\
- Instruction intent misunderstood or disregarded\\
- Caption appears written without considering the instruction\\
\#\#\# Instruction-Following Accuracy: 1--10 Scale\\
**9-10 (Exceptional)**: Near-perfect instruction adherence. All explicit and implicit requirements met with natural, high-quality execution. Any deviations are trivial.\\
**7-8 (Strong)**: Excellent instruction-following with minor gaps. All critical requirements met. May miss 1-2 minor elements or execute some aspects imperfectly, but overall very aligned.\\
**5-6 (Adequate)**: Satisfactory instruction-following with notable gaps. Core requirements met but several secondary elements missed or poorly executed. Instruction intent generally honored.\\
**3-4 (Weak)**: Poor instruction-following with major gaps. Many requirements missed or violated. Some alignment with instruction but significant divergence in execution.\\
**1-2 (Very Weak)**: Minimal instruction-following. Most requirements ignored. Caption appears largely independent of the instruction. Major misunderstanding or disregard of requirements.\\
---\\
\#\# Special Considerations\\
\#\#\# Edge Cases to Watch For:\\
1. **Conflicting Requirements**: If instruction contains contradictory elements, evaluate based on how well the model navigates the conflict.\\
2. **Ambiguous Instructions**: If instruction is vague, give credit for reasonable interpretations that align with common sense.\\
3. **Impossible Requirements**: If instruction asks for information not determinable from audio alone, don't penalize the model for not hallucinating content.\\
4. **Over-compliance**: If model follows instruction so rigidly that caption quality suffers dramatically, note this in justification but still credit the instruction-following attempt.\\
5. **Creative Interpretation**: Distinguish between helpful creative interpretation (good) and ignoring instructions (bad).\\
\#\#\# Common Pitfalls to Identify:\\
- ❌ **Instruction ignored entirely**: Caption is generic and could apply to any instruction\\
- ❌ **Partial attention**: Model addresses only the first part of a multi-part instruction\\
- ❌ **Format violation**: Instruction specifies structure but caption uses different format\\
- ❌ **Tone mismatch**: Formal instruction produces casual output, or vice versa\\
- ❌ **Scope creep**: Model includes content explicitly excluded by instruction\\
- ❌ **Misinterpretation**: Model misunderstands key terms or intent of instruction\\
---\\
\#\# Output Format\\
Provide your evaluation in the following structure:\\
\textless analysis\textgreater\ [Identify the key requirements in the instruction: What are the explicit demands? What are the implicit expectations? What constraints are specified? Categorize them as content, structural, stylistic, focus, or constraint requirements.]\ \textless/analysis\textgreater\\
\textless compliance\_assessment\textgreater\ [For each identified requirement, evaluate whether the caption meets it. Mark each as: ✓ Met, ⚠ Partially Met, or ✗ Not Met. Provide brief evidence from the caption.]\ \textless/compliance\_assessment\textgreater\\
\textless strengths\textgreater\ [Highlight what the model did well in following the instruction. What requirements were executed particularly effectively?]\ \textless/strengths\textgreater\\
\textless weaknesses\textgreater\ [Identify gaps, violations, or missed opportunities. What requirements were ignored, poorly executed, or misinterpreted?]\ \textless/weaknesses\textgreater\\
\textless justification\textgreater\ [Synthesize your analysis into a coherent evaluation. Explain your success determination and accuracy score. Reference specific examples from both the instruction and caption. Consider the severity of any violations and the overall alignment with instruction intent.]\ \textless/justification\textgreater\\
\textless ratings\textgreater\\
\{ "instruction\_following\_success": "[Success/Partial Success/Failure]", "instruction\_following\_accuracy": [1-10] \}\\
\textless/ratings\textgreater\\
---\\
\#\# Important Reminders\\
1. **Focus on instruction compliance, not caption quality**: A poorly written caption that follows instructions perfectly should score higher than a beautifully written caption that ignores them.\\
2. **Be precise in your reasoning**: Cite specific phrases from both the instruction and caption to support your evaluation.\\
3. **Distinguish between critical and minor requirements**: Not all instruction elements carry equal weight. Prioritize core requirements over secondary details.\\
4. **Consider context**: Audio captioning has domain-specific norms. Evaluate within the context of speaking-style description conventions.\\
5. **Be fair but rigorous**: Give credit for good-faith attempts while maintaining high standards for what constitutes successful instruction-following.\\
6. **Avoid lenience bias**: Don't give full credit just because the caption is reasonable. It must specifically follow the given instruction.\\
Begin your evaluation now.
\end{minipage}
};
\end{tikzpicture}
\caption{Instruction-following evaluation prompt (Part 2): scoring rubric, edge cases, and required output format.}
\label{fig:prompt-instruction-following-eval-b}
\end{figure*}

\begin{figure*}[t]
\centering
\begin{tikzpicture}
\node[
  draw,
  rounded corners=2pt,
  line width=0.4pt,
  inner sep=5pt,
  text width=0.97\textwidth,
  align=left
] (bank) {%
\begin{minipage}{\linewidth}
\ttfamily\scriptsize
\setlength{\parindent}{0pt}
\setlength{\parskip}{0pt}
\setlength{\baselineskip}{0.93\baselineskip}

\{\\
\ \ "unseen\_prompt\_1": "Establish the speaker’s vocal identity without mentioning or inferring gender. Use only acoustic cues to describe pitch range, timbre/texture, resonance, and any stable voice qualities; if a trait is uncertain, state it as unknown rather than guessing.",\\
\ \ "unseen\_prompt\_2": "Write a speaker-identity caption that avoids any accent or origin claims. Focus on voice quality (e.g., breathiness/creak/clarity), pitch behavior, articulation style, and timbral color, relying strictly on what can be heard.",\\
\ \ "unseen\_prompt\_3": "Output a Python-dict-style profile (not JSON) with exactly these keys: \{\"pitch\_character\": ..., \"timbre\": ..., \"vocal\_texture\": ..., \"stability\": ..., \"notes\": ...\}. Fill values with short phrases grounded in the audio; use null for anything you cannot justify.",\\
\ \ "unseen\_prompt\_4": "Provide two competing hypotheses for the speaker’s vocal identity (Hypothesis A vs B) based only on acoustics, then pick the better one with a confidence score from 0 to 1. Do not mention language, name, or demographics.",\\
\ \ "unseen\_prompt\_5": "Describe the speaker as a casting director evaluating a voice actor for a role. Keep it strictly evidence-based: every descriptive claim must correspond to an audible cue (pitch movement, timbre, articulation, resonance).",\\
\ \ "unseen\_prompt\_6": "Produce a single sentence that captures the speaker’s ‘voiceprint’ using neutral phrasing and no demographic attributes. The sentence must include exactly one timbre descriptor, one pitch descriptor, and one articulation descriptor.",\\[0.25em]

\ \ "unseen\_prompt\_7": "Infer the recording context (environment + microphone distance) and present it as: (1) best-guess setting, (2) 2-3 acoustic cues supporting it, (3) an uncertainty note. Do not mention emotion or speaker identity.",\\
\ \ "unseen\_prompt\_8": "Write a ‘field recording log’ in three bullet points: Background (noise type + stability), Channel (reverb/roominess + clipping/cleanliness), Distance (near/far + evidence). Avoid guessing location names.",\\
\ \ "unseen\_prompt\_9": "Return a compact YAML block with keys: setting\_guess, noise\_profile, reverberation, mic\_distance, confidence. Values must be short, and confidence must be one of: low/medium/high.",\\
\ \ "unseen\_prompt\_10": "Describe how the environment changes over time across the utterance (start → middle → end). Only reference audible changes (noise bursts, reverb shifts, level changes); do not hallucinate events.",\\
\ \ "unseen\_prompt\_11": "As an audio engineer, diagnose whether the speech sounds studio-like or in-the-wild and justify your conclusion with at least three acoustic indicators. If evidence is mixed, explicitly say so.",\\
\ \ "unseen\_prompt\_12": "Give a one-sentence context caption that must avoid these words: room, background, noise, microphone. You may still describe context using alternative phrasing grounded in acoustics.",\\[0.25em]

\ \ "unseen\_prompt\_13": "Describe the affect using only valence (1-7) and arousal (1-7), plus short acoustic evidence (pitch dynamics, intensity, tempo, voice quality). Do not use categorical emotion labels.",\\
\ \ "unseen\_prompt\_14": "Return a strict JSON object with exactly these keys: \{\"valence\":1-7, \"arousal\":1-7, \"confidence\":0-1, \"evidence\":[...]\}.",\\
\ \ "unseen\_prompt\_15": "Write two sentences: Sentence 1 must be purely objective (acoustic description). Sentence 2 may interpret the likely feeling state but must include a caution clause about uncertainty.",\\
\ \ "unseen\_prompt\_16": "Describe the emotional trajectory across the utterance and name the specific prosodic features that signal each phase. Avoid any mention of the textual content.",\\
\ \ "unseen\_prompt\_17": "As a clinician writing a neutral intake note, summarize the speaker’s affective presentation using cautious language and explicit acoustic grounding; do not overclaim.",\\
\ \ "unseen\_prompt\_18": "Generate a caption that includes exactly one metaphor for the affect, immediately followed by concrete acoustic justification.",\\[0.25em]

\ \ "unseen\_prompt\_19": "Infer the likely communicative intent from prosody alone and provide 2-3 prosodic cues as justification.",\\
\ \ "unseen\_prompt\_20": "Output a ranked list of up to three pragmatic functions with probability estimates that sum to 1.0.",\\
\ \ "unseen\_prompt\_21": "Write as a debate judge assessing delivery: identify confidence vs hesitation and strategic emphasis.",\\
\ \ "unseen\_prompt\_22": "Produce a minimal-pair analysis contrasting the most likely intent with a plausible alternative.",\\
\ \ "unseen\_prompt\_23": "Return a markdown table: Prosodic cue | What it signals | Confidence (low/med/high).",\\
\ \ "unseen\_prompt\_24": "Create a single-sentence pragmatic caption that must not include any pitch words.",\\[0.25em]

\ \ "unseen\_prompt\_25": "Rate warmth, urgency, and confidence (1-5) with one acoustic justification per scale.",\\
\ \ "unseen\_prompt\_26": "As a call-center QA evaluator, describe perceived clarity and assertiveness, separating observation from interpretation.",\\
\ \ "unseen\_prompt\_27": "Write a listener-centric caption focusing only on intelligibility and comfort.",\\
\ \ "unseen\_prompt\_28": "Return a strict JSON object with keys: impression, accessibility\_risks, most\_salient\_cue, confidence.",\\
\ \ "unseen\_prompt\_29": "Advise a TTS system how to match the perceived presence and approachability using audible cues only.",\\
\ \ "unseen\_prompt\_30": "Give two captions: one for a casual listener and one for a trained listener with technical descriptors.",\\[0.25em]

\ \ "unseen\_prompt\_31": "Write director’s notes for recreating this performance, grounded strictly in acoustics.",\\
\ \ "unseen\_prompt\_32": "Produce two labeled lines: FACT and IMAGERY, ensuring IMAGERY does not contradict FACT.",\\
\ \ "unseen\_prompt\_33": "Return a nested JSON object summarizing speaker, delivery, and context; use null for unknowns.",\\
\ \ "unseen\_prompt\_34": "Write a sports-commentator-style description justified by explicit acoustic cues.",\\
\ \ "unseen\_prompt\_35": "Create a single sentence of exactly 18 words integrating identity, delivery, and context.",\\
\ \ "unseen\_prompt\_36": "Write a three-bullet spec sheet (Identity / Delivery / Context) with acoustic justification."\\
\}\\
\end{minipage}
};
\end{tikzpicture}
\caption{Unseen instruction prompts used to probe diverse instruction-following behaviors for speaking-style captioning.}
\label{fig:unseen-instruction-prompts}
\end{figure*}

\newpage


\newcommand{\JustificationText}{\detokenize{
The generated caption matches the ground truth in gender, accent, pitch, and speaking rate. It captures the emotional tone of discomfort and frustration, though it's slightly less specific than the ground truth's pained emotion. The generated caption introduces a nasal voice quality not mentioned in the ground truth, which is acceptable as per instructions. The environment/noise level is missing, which is a notable difference.
}}

\newcommand{\GroundTruthMeta}{\detokenize{
{
  "source": "ears",
  "relative_audio_path": "p019/emo_pain_freeform.wav",
  "text_description": [
    " A male speaker with an American accent delivers a speech that is deep and high-pitched, characterized by crisp, husky tones. His voice is measured in speed, punctuated with slight pauses, conveying a sense of pained emotion. The speech was recorded in a slightly noisy environment, adding a flowing, textured quality to the overall sound.",
    " A male speaker with a high-pitched voice delivers pained expressions in an American accent, speaking at a measured speed in a slightly noisy environment."
  ],
  "transcription": " Ow, god damn it, that hurts. Damn, I wish I hadn't stubbed my toe on the corner of that table. Oh, that's kind of smart. Oh, boy, I hate that so much. Why did it have to be the pinky toe as well? Ow.",
  "intrinsic_tags": ["crisp", "deep", "american", "flowing", "husky", "punctuated"],
  "situational_tags": ["pained"],
  "basic_tags": ["high-pitched", "male", "measured speed", "slightly noisy environment"],
  "all_tags": ["american", "crisp", "deep", "flowing", "high-pitched", "husky", "male", "measured speed", "pained", "punctuated", "slightly noisy environment"],
  "speakerid": "p019",
  "name": "p019",
  "duration": 16.1,
  "gender": "male",
  "accent": "american",
  "pitch": "high-pitched",
  "speaking_rate": "measured speed",
  "noise": "slightly noisy environment",
  "utterance_pitch_mean": 159.45840454101562,
  "snr": 38.82289123535156,
  "holistic_creative_synthesis": "A high-pitched, crisp, and husky American male voice speaks in a slightly noisy environment, conveying a sense of pained emotion through a measured pace and deliberate pauses, the speaker's words punctuated by a rhythmic flow that adds to the overall sense of conviction, creating a sense of depth and nuance that draws the listener in.",
  "holistic_creative_synthesis_instruction": "Combine all elements of the speaker's style, including their accent, pitch, timbre, and delivery, to create a rich, narrative-style caption that captures the essence of their speaking-style."
}
}}

\newcommand{\GeneratedCaptionMeta}{\detokenize{
Generated caption: The speaker's style is characterized by a low, guttural tone, a slow and deliberate pace, and a deep, resonant timbre. Their accent is distinct and evocative, with a strong emphasis on certain syllables and a distinctive rhythm. The delivery is confident and assertive, with a sense of authority and expertise. Overall, the speaker's style is engaging and captivating, drawing the listener in with their unique voice and delivery.
}}

\twocolumn[{
\section{Model Training Details}
\label{app:model_training}
\vspace{-2mm}

\centering
\small
\setlength{\tabcolsep}{4pt}
\renewcommand{\arraystretch}{1.05}
\urlstyle{same}
\expandafter\def\expandafter\UrlBreaks\expandafter{\UrlBreaks\do\-}

\begin{tabularx}{\textwidth}{@{}l l >{\raggedright\arraybackslash}X@{}}
\toprule
\textbf{Models} & \textbf{Role} & \textbf{Sources} \\
\midrule
Audio-Flamingo-3        & Evaluated & \url{https://huggingface.co/nvidia/audio-flamingo-3} \\
Voxtral-Mini-3B         & Evaluated & \url{https://huggingface.co/mistralai/Voxtral-Mini-3B-2507} \\
Voxtral-Small-24B       & Evaluated & \url{https://huggingface.co/mistralai/Voxtral-Small-24B-2507} \\
Qwen2-Audio-7B-Instruct & Evaluated & \url{https://huggingface.co/Qwen/Qwen2-Audio-7B-Instruct} \\
MERaLiON-AudioLLM-Whisper-SEA-LION          & Evaluated & \url{https://huggingface.co/MERaLiON/MERaLiON-AudioLLM-Whisper-SEA-LION} \\

Qwen3-Omni & Evaluated & \url{https://huggingface.co/Qwen/Qwen3-Omni-30B-A3B-Instruct} \\
MOSS-Audio & Evaluated & \url{https://github.com/OpenMOSS/MOSS-Audio} \\
Audio-Flamingo-Next & Evaluated & \url{https://huggingface.co/nvidia/audio-flamingo-next-hf} \\

SALMONN-7B             & Evaluated & \url{https://github.com/bytedance/SALMONN/tree/salmonn} \\
OpenAI-GPT-Audio & Evaluated (closed-source reference) & \url{https://platform.openai.com/docs/models/gpt-audio} \\
\midrule
Mistral-Small-24B-Instruct & Caption generation for train/validation & \url{https://huggingface.co/mistralai/Mistral-Small-24B-Instruct-2501} \\
Llama-3.1-8B-Instruct      & Caption generation for evaluation splits & \url{https://huggingface.co/meta-llama/Llama-3.1-8B-Instruct} \\

Gemma-2-27B      & Caption generation for adversarial caption set & \url{https://huggingface.co/google/gemma-2-27b} \\
Qwen-3-32B                 & LLM-Judge & \url{https://huggingface.co/Qwen/Qwen3-32B} \\
Gemma-4-31B-it & LLM-Judge & \url{https://huggingface.co/google/gemma-4-31B-it} \\
GPT-OSS-20B & LLM-Judge & \url{https://huggingface.co/openai/gpt-oss-20b} \\
\bottomrule
\end{tabularx}

\captionof{table}{Specification and sources of the employed models.}
\label{table:models}

\vspace{2mm}

}]

\paragraph{Training and inference details.}
We fine-tune the SALMONN variants using LoRA to control adaptation capacity and limit task-specific collapse. We set the LoRA rank to $r{=}8$ throughout. During training, we use a stronger scaling ($\alpha{=}32$) to enable rapid alignment to the instruction-following format and the speech-conditioning interface; at inference, we reduce the scaling to $\alpha{=}16$. This inference-time attenuation acts as a practical ``de-overfitting'' knob: it dampens overly task-specialized update directions while preserving the base LLM's general generation ability, improving robustness to prompt and domain shifts. Empirically, this reduces random stylistic drift (e.g., ungrounded claims about affect or speaker traits) and encourages generations to remain anchored to the acoustic evidence rather than the most frequent training-task patterns.

\paragraph{Instruction-tuning regimes and batch mixing.}
We study two instruction-tuning regimes. \textbf{Single-task} tuning introduces one new task, which isolates the effect of task supervision but also makes memorization easier. \textbf{Multi-task} tuning introduces six new tasks, where tasks are sampled uniformly at random within each mini-batch. This stochastic task mixture increases gradient diversity and discourages the model from specializing to any one instruction template. We additionally explored structured batch sampling strategies (e.g., round-robin task cycling and scheduled curricula), but these did not improve performance and in several cases degraded it, likely because deterministic mixing reduces the regularizing effect of interleaving and amplifies short-horizon task dominance within contiguous training windows.

\paragraph{Scaling data and diagnosing task overfitting.}
Training on the PSC-base human-annotated subset (282 hours) yields a consistent failure mode: (i) the model overfits to the limited set of training tasks, (ii) it produces less expressive captions (tending toward safe, generic phrasing), and (iii) it exhibits catastrophic forgetting, struggling to answer open-set questions that were not explicitly present in the tuning set. We interpret this as a distributional ``compression'' effect: with limited diversity, the model learns a narrow mapping from acoustics to a small set of instruction-conditioned outputs, trading off nuanced style description for high-confidence task compliance. Increasing scale mitigates these issues: training on 100\% of PSC and expanding instruction diversity to roughly $\sim$300 prompts per task substantially improves caption quality and stylistic richness. However, the underlying task-overfitting tendency persists, suggesting that capacity control and objective design (not only data volume) are necessary to fully address specialization.

\paragraph{Alternative learning strategies for style grounding.}
To improve grounding and reduce spurious style attributions, we investigate two complementary strategies.
(\textbf{A}) \textbf{Auxiliary-task inclusion:} we incorporate automatic speech recognition (ASR), gender recognition (GR), and accent recognition (AR) as additional supervised objectives. These tasks encourage the shared representation to encode linguistically and paralinguistically salient cues, acting as an inductive bias toward acoustically-verifiable factors.
(\textbf{B}) \textbf{Factor-conditioned captioning:} we modify the training target to first predict explicit factors (e.g., \texttt{gender}, \texttt{accent}, \texttt{pitch}, \texttt{speaking rate}) followed by the free-form speaking-style caption. By forcing the model to commit to concrete attributes before generating the descriptive caption, we aim to reduce shortcut learning and improve causal alignment between acoustics and the final narrative description.

\paragraph{Transcript as linguistic context.}
We provide the transcript as linguistic context for speaking-style captioning. This yields a complementary perspective: it allows the model to separate \emph{what} was said (lexical content) from \emph{how} it was said (prosody, voice quality, pacing), and to interpret paralinguistic signals relative to discourse intent (e.g., emphasis, question intonation, sarcasm markers). To our knowledge, explicitly incorporating the transcript in this way has not been standard in prior speaking-style captioning setups, and it improves the model's ability to generate contextually appropriate style descriptions rather than purely acoustic correlates.

\paragraph{Decoding.}
At inference time, we use default decoding settings for all models unless otherwise noted. For SALMONN, we use deterministic beam search to prioritize stability and reduce sampling-induced variance:
\texttt{max\_new\_tokens}{=}200,
\texttt{num\_beams}{=}4,
\texttt{do\_sample}{=}\texttt{False},
\texttt{min\_length}{=}1,
\texttt{temperature}{=}1.0,
\texttt{top\_p}{=}0.9,
\texttt{repetition\_penalty}{=}1.0,
\texttt{length\_penalty}{=}1.0.

\FloatBarrier

\section{Lexical Diversity}
\label{app:lexical-diversity}
\vspace{-2mm}
\begin{table*}[!t]
\centering
\small
\setlength{\tabcolsep}{6pt}
\begin{tabular}{lcccccc}
\toprule
\textbf{Caption set} &
\textbf{HD-D} &
\textbf{Self-BLEU} &
\textbf{Distinct-2} &
\textbf{Distinct-3} &
\textbf{\makecell{Mean caption\\length}} &
\textbf{\makecell{STD of\\caption lengths}} \\
\midrule

Oracle caption (PSC)         & 0.732 & ---    & 0.0425 & 0.0795 & 32.5 & 7.97 \\
Oracle caption (SIC)         & 0.771 & 0.304  & 0.152  & 0.310  & 34.9 & 12.5 \\

Audio-Flamingo-3             & 0.748 & 0.0881 & 0.138  & 0.248  & 31.8 & 23.0 \\
Voxtral-Mini-3B              & 0.813 & 0.123  & 0.247  & 0.463  & 145  & 78.1 \\
Voxtral-Small-24B            & 0.807 & 0.145  & 0.281  & 0.484  & 72.0 & 56.8 \\
Qwen2-Audio-7B-Instruct      & 0.836 & 0.0362 & 0.231  & 0.439  & 86.2 & 59.5 \\
MERaLiON-AudioLLM            & 0.782 & 0.0826 & 0.259  & 0.429  & 29.9 & 21.6 \\

MOSS-Audio                   & 0.836 & 0.0478 & 0.342  & 0.631  & 56.6 & 26.4 \\

Qwen3-omni                   & 0.856 & 0.0641 & 0.343  & 0.610  & 134  & 92.0 \\

Audio-Flamingo-next          & 0.885 & 0.128  & 0.330  & 0.501  & 46.8 & 44.4 \\

SALMONN                      & 0.795 & 0.113  & 0.221  & 0.397  & 84.3 & 32.4 \\

SALMONN (PSC)                & 0.735 & 0.361  & 0.0906 & 0.195  & 66.4 & 18.9 \\
SALMONN (SIC-DB) - ST        & 0.786 & 0.315  & 0.135  & 0.271  & 72.8 & 15.4 \\
SALMONN (SIC-DB) - MT        & 0.776 & 0.206  & 0.137  & 0.266  & 52.8 & 18.0 \\

\bottomrule
\end{tabular}
\caption{Lexical diversity and caption-length statistics across different caption sets and models.}
\label{tab:caption_diversity_stats}
\end{table*}

From the lexical diversity analysis in Table \ref{tab:caption_diversity_stats}, we find that task-specific fine-tuning generally reduces linguistic diversity, which is expected since the model is optimized to consistently express a constrained set of speaking-style attributes rather than maximize unconstrained lexical variation. Compared with single-task training, multi-task training improves faithfulness and reduces hallucinations (Table~\ref{tab:LLM_judge_3_evals}), but its effect on diversity is mixed: it clearly reduces repetition across captions (Self-BLEU 0.206 vs.\ 0.315), whereas HD-D and Distinct-3 are slightly lower and Distinct-2 is essentially unchanged. A larger LLM backbone also does not consistently increase diversity. From Voxtral-Mini-3B to Voxtral-Small-24B, Distinct-2 and Distinct-3 increase, but HD-D decreases slightly and Self-BLEU rises, indicating more repetition. We additionally observe that the oracle captions in StyleInstructCaps are considerably more diverse than those in ParaSpeechCaps. Models trained on StyleInstructCaps are not uniformly more diverse than the caption set itself. Both SALMONN (SIC-DB) variants exceed the oracle captions on HD-D, and the MT variant has lower Self-BLEU, but both have lower Distinct-2 and Distinct-3.

\FloatBarrier
\clearpage
\twocolumn[{
\section{LLM Judgement}
\vspace{-2mm}
\label{app:LLM_judgement}
\centering
\begin{tikzpicture}[
  every node/.style={font=\ttfamily},
  panel/.style={
    draw=black!55,
    line width=0.35pt,
    rounded corners=1.6mm,
    fill=black!2,
    align=left,
    inner xsep=10pt,
    inner ysep=9pt
  },
  rule/.style={draw=black!35, line width=0.35pt},
]

\node[panel, text width=0.965\linewidth] (judge) {%
  {\hdr LLM judgement}\par
  \vspace{-2pt}\tikz\draw[rule] (0,0)--(\linewidth,0);\par\vspace{3pt}
  \textbf{similarity\_score:} 7\par\vspace{3pt}
  \textbf{justification:}\par\vspace{3pt}
  {\scriptsize\linespread{0.85}\selectfont\JustificationText\par}
};

\node[panel, text width=0.5\linewidth, below=7mm of judge.south west, anchor=north west] (gt) {%
  {\hdr Ground truth metadata}\par
  \vspace{-2pt}\tikz\draw[rule] (0,0)--(\linewidth,0);\par\vspace{3pt}
  {\scriptsize\linespread{0.85}\selectfont\GroundTruthMeta\par}
};

\node[panel, text width=0.4\linewidth, below=7mm of judge.south east, anchor=north east] (gen) {%
  {\hdr Generated caption}\par
  \vspace{-2pt}\tikz\draw[rule] (0,0)--(\linewidth,0);\par\vspace{3pt}
  {\scriptsize\linespread{0.85}\selectfont\GeneratedCaptionMeta\par}
};

\end{tikzpicture}

\captionof{figure}{Example showing an LLM judgment (top) and the corresponding ground-truth metadata vs.\ generated caption (bottom).}
\label{fig:judge_record_p019_pretty}
\vspace{2mm}
}]

\FloatBarrier
\clearpage








\FloatBarrier
\clearpage

\twocolumn[{
\section{Speaker Caption clustering plots}
\label{app:spk_cluster}

\centering
\includegraphics[draft=false,width=\textwidth,height=0.9\textheight,keepaspectratio]{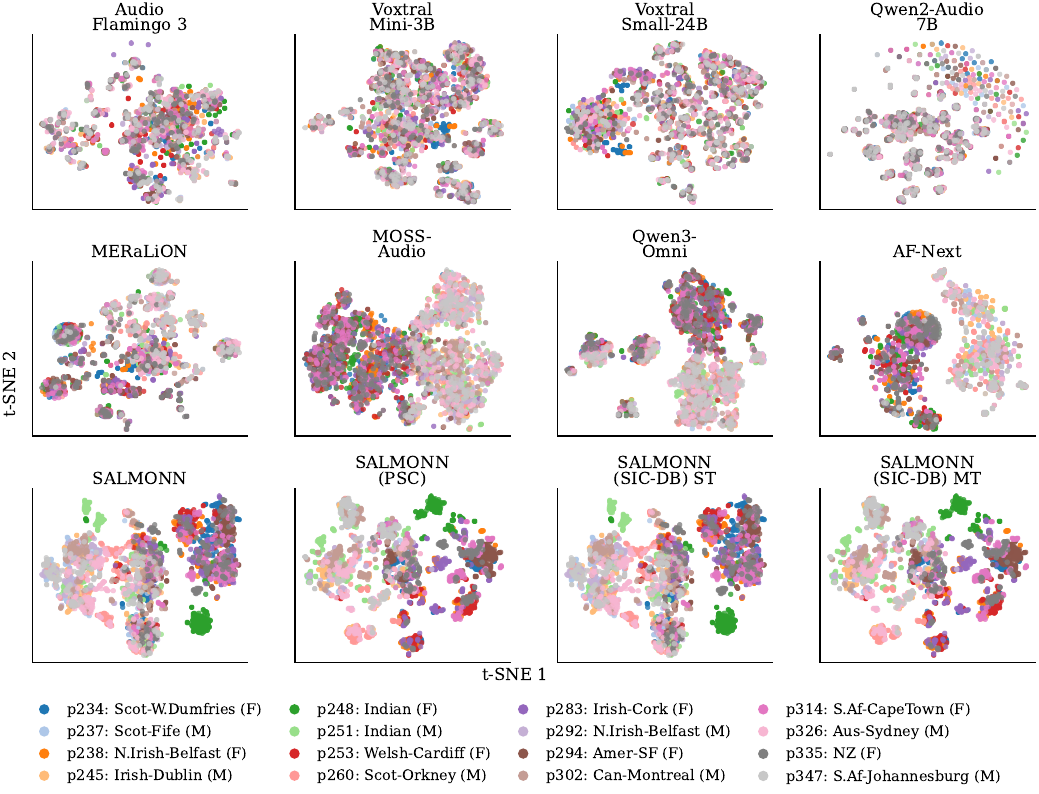}
\captionof{figure}{t-SNE visualization of speaker-style caption
embeddings on UD-VCTK (all speakers)}
\label{fig:tsne_allspks_all_models}

\vspace{2mm} 
}]

\FloatBarrier
\clearpage

\section{Use of AI assistants}
We used GPT-5 and Gemini 3 Flash for code assistance, and the same models purely for language-related assistance in writing the paper.

\section{Licensing}
All resources used in this work are released under open licenses. We use the VCTK corpus, which is distributed under the Open Data Commons Attribution License (ODC-By) v1.0, requiring attribution. 
We also use the GigaSpeech dataset, released under the Apache License 2.0. 
ParaSpeechCaps is provided under the CC BY-NC-SA 4.0 license, which restricts commercial use and requires attribution and share-alike for derivatives.
Finally, the SALMONN codebase is released under the Apache License 2.0. 


\end{document}